\documentclass{article}

\PassOptionsToPackage{numbers,sort&compress}{natbib}
\usepackage{genai4health_2026}

\usepackage{iftex}
\ifPDFTeX
  \usepackage[utf8]{inputenc}
\fi
\usepackage[T1]{fontenc}
\usepackage{hyperref}
\usepackage{url}
\usepackage{booktabs}
\usepackage{array}
\usepackage{amsfonts}
\usepackage{wrapfig}
\usepackage{amsmath}
\usepackage{amssymb}
\usepackage{mathtools}
\usepackage{subcaption}
\usepackage{nicefrac}
\usepackage{microtype}
\usepackage{multirow}
\usepackage[dvipsnames]{xcolor}
\usepackage{graphicx}
\usepackage{float}
\newcommand{\fullskips}{\setlength{\abovedisplayshortskip}{\abovedisplayskip}\setlength{\belowdisplayshortskip}{\belowdisplayskip}}

\hypersetup{
    colorlinks=true,
    citecolor=blue,
    linkcolor=black,
    urlcolor=blue,
    pdftitle={Electrocardiographic Imaging via Posterior Sampling under Operator Discrepancy},
    pdfauthor={Ricarda Buttmann and Andreas M. Lehrmann},
}

\title{Electrocardiographic Imaging via\\Posterior Sampling under Operator Discrepancy}

\author{%
  Ricarda Buttmann and Andreas M. Lehrmann \\
  Center for Artificial Intelligence and Robotics \\
  Technical University of Applied Sciences \\
  Würzburg-Schweinfurt, Germany \\
  \texttt{\{ricarda.buttmann,andreas.lehrmann\}@thws.de}
}

\begin{document}

\maketitle

\begin{abstract}
Electrocardiographic imaging (ECGI) reconstructs heart-surface potentials from body-surface potentials, an ill-posed inverse problem. Diffusion posterior samplers combine a learned prior with an explicit forward operator that they assume to be accurate. In ECGI this operator is only an estimate, because the body changes with respiration and posture, and body composition such as subcutaneous fat is rarely known for an individual patient. The resulting operator discrepancy is structured and depends on the heart signal. We introduce OpDis-PS, a diffusion posterior sampler that models this discrepancy instead of treating it as signal-independent noise. OpDis-PS learns how the discrepancy depends on the signal, folds this into a corrected forward operator, and reconstructs with a fixed prior across all operator conditions. We study two realistic sources of discrepancy, changing lung conductivity and an added subcutaneous-fat layer. Across both, OpDis-PS reconstructs more accurately than classical regularization and learned predictors, with its largest gains on held-out cardiac pathologies.
\end{abstract}

\section{Introduction}
\label{sec:introduction}

Cardiovascular diseases remain the leading cause of mortality worldwide, accounting for an estimated 19.8 million deaths in 2022 \citep{Mensah2023GBD}. The diagnosis and treatment of arrhythmias and conduction disorders often rely on identifying where abnormal ventricular activation originates and how it propagates. Electrocardiographic imaging (ECGI) reconstructs heart-surface potentials (HSPs), defined here on the outer surface of the ventricles, from body-surface potentials (BSPs) measured on the torso \citep{RudyMessingerRapport1988}. These spatio-temporal signals reveal ventricular activation and recovery without requiring electrodes on the cardiac surface and can support the analysis and treatment of arrhythmias \citep{Cluitmans2018}. However, the torso attenuates and spatially smooths the cardiac field \citep{Ramanathan2001}, rendering ECGI an ill-conditioned inverse problem that amplifies measurement and model errors.

Traditional ECGI methods stabilize the inversion using Tikhonov penalties, spatio-temporal constraints, or total variation \citep{MessingerRapportRudy1988MathBiosci,Brooks1999,RudinOsherFatemi1992}. While these regularizers improve stability, their generic smoothness assumptions capture only limited physiological structure and can attenuate localized activation patterns. Learning-based methods can capture complex cardiac dynamics directly from data, with approaches ranging from discriminative predictors that estimate HSPs directly from BSPs to conditional generative models that learn the distribution of HSPs given BSP observations~\citep{Bacoyannis2021,JaraMeyers2026}. The latter are particularly suited to representing uncertainty and ambiguity in the inverse solution. Despite this advantage, both approaches encode the observation process in their parameterization and do not permit substituting an explicit case-specific forward operator at inference. 

An alternative is to learn a generative prior over cardiac signals independently of the observation model and combine the two only during inference. Diffusion posterior samplers such as Denoising Diffusion Restoration Models (DDRM) and Diffusion Posterior Sampling (DPS) realize this separation by coupling a learned diffusion prior with an explicit measurement-consistency model \citep{Kawar2022,Chung2023}. Originally introduced for vision tasks, both methods assume that the supplied forward operator accurately represents the measurement process. In ECGI, however, the nominal operator used for reconstruction is only an estimate derived from an anatomical model and assumed tissue conductivities. The recording operator, which represents the actual volume conductor at measurement time, can differ from this nominal operator because of respiration, posture, electrode placement, or body composition. The resulting operator discrepancy produces a structured error that depends on the underlying cardiac signal. Because ECGI is ill-conditioned, even a modest operator discrepancy can cause substantial reconstruction errors.

We propose OpDis-PS, a diffusion posterior-sampling method for reconstruction under forward-operator discrepancy. Rather than treating operator discrepancy as signal-independent measurement noise, OpDis-PS models its statistical dependence on the nominal forward signal. A signal-conditional Gaussian approximation-error model yields a linear correction to the nominal operator and a covariance for the remaining mismatch. We estimate these quantities offline using HSPs from the training set paired with sampled recording operators. At reconstruction time, OpDis-PS combines the residual discrepancy covariance with the measurement-noise covariance and whitens the corrected observation model before applying a spectral posterior sampler. The unconditional diffusion prior remains fixed across operator conditions, allowing the likelihood to adapt without retraining the generative model.\footnote{Code, pretrained models, and datasets: \href{https://projects.cairo.thws.de/asml/papers/ecgi/}{projects.cairo.thws.de/asml/papers/ecgi}.} We evaluate OpDis-PS under two physiologically motivated sources of operator discrepancy: lung-conductivity variation, representing different lung inflation states \citep{Nopp1993,Keller2010}, and a superficial subcutaneous-fat layer, representing a body-composition-related torso inhomogeneity known to affect BSPs \citep{Klepfer1997,Bradley2000}. Across these settings, OpDis-PS is compared with baselines spanning multiple learning paradigms and demonstrates strong reconstruction performance.

\textbf{Contributions.} In summary, we make the following contributions: ({\bf 1}) we present OpDis-PS, a diffusion posterior sampler that reconstructs heart-surface potentials when the reconstruction operator differs from the one that generated the measurement; ({\bf 2}) we show how to cast this operator discrepancy as a signal-conditional approximation-error term, absorb its predictable part into a corrected operator and observation, and carry the residual uncertainty into a whitened data-consistency step that leaves the diffusion prior untouched; and ({\bf 3}) we build a controlled ECGI evaluation from openCARP \citep{Plank2021openCARP} simulations that isolates operator discrepancy from measurement noise, and assess OpDis-PS against variational, amortized, and posterior-sampling methods on physiological and unseen pathological activation.

\section{Related work}
\label{sec:related-work}

OpDis-PS relates to four lines of prior work. We discuss foundational ECGI studies, classical regularizers, machine-learning models for ECGI, and the diffusion posterior samplers on which OpDis-PS builds.

{\bf General ECGI problem.}
Early studies establish the physical and mathematical setting of ECGI. \citet{RudyMessingerRapport1988} formulate the recovery of HSPs from BSPs as a linear inverse problem and analyze its ill-posed structure. \citet{Oster1997} subsequently demonstrate such reconstructions in a torso-tank experiment. \citet{Ramanathan2001} study the forward problem and examine how volume conduction and torso inhomogeneities affect BSPs.

{\bf Traditional methods.}
Classical ECGI reconstruction relies on variational regularizers that penalize spatial or temporal features of the HSPs. Comparative studies examine Total Variation and first- and second-order Tikhonov regularization among other inverse approaches \citep{Milanic2014,Karoui2018}. \citet{Cluitmans2017} replace generic penalties with a truncated basis learned from simulated activation and recovery patterns and reconstruct sparse coefficients in this basis. Moving from deterministic regularization to a probabilistic treatment of forward-operator error, \citet{AldemirDogrusoz2021} model the approximation error as an additive Gaussian term with a data-estimated covariance and incorporate it into a Bayesian MAP reconstruction. OpDis-PS carries this approximation-error idea into diffusion posterior sampling and extends it with a signal-conditional operator correction learned from paired training data.

{\bf Machine learning methods for ECGI.}
Data-driven ECGI methods differ in their reconstruction targets, treatment of geometry, and use of the forward operator. Amortized end-to-end approaches train networks that map BSPs directly to heart-surface quantities without explicitly using the forward operator at inference \citep{Bujnarowski2020,Jiang2023,Wang2025}. \citet{JaraMeyers2026} use conditional diffusion to map BSPs to HSPs without subject-specific geometry. Our Transformer and Palette \citep{Saharia2022Palette} baselines represent this amortized-inference line and operate on a common heart-template representation across subjects. Other generative approaches use latent-variable models: \citet{Ghimire2018} reconstruct volumetric transmembrane-potential sequences using a subject-specific sequential VAE prior, while \citet{Bacoyannis2021} use a conditional VAE to sample volumetric activation-time maps.

{\bf Diffusion posterior samplers.}
\citet{Chung2023} propose DPS, which guides diffusion sampling using a gradient of measurement consistency. \citet{Kawar2022} formulate DDRM in the singular coordinates of a linear forward operator, with updates determined by the singular values and measurement-noise level. \citet{Nezakati2026CARD} extend DDRM to correlated Gaussian measurement noise by whitening the observation before the restoration update. These approaches formulate data consistency using a single observation operator that is assumed to be accurate at inference. OpDis-PS relaxes this assumption by combining a signal-conditional correction of the nominal operator with a residual discrepancy model in the whitened likelihood.

\section{Technical prerequisites}\label{sec:background}

The ECGI forward process propagates heart-surface potentials through a torso volume conductor to body-surface potentials. OpDis-PS builds upon previous work on ({\bf 1}) approximating this continuous process as a linear model with Gaussian measurement noise (Section~\ref{sec:background-forward}) and ({\bf 2}) inverting linear models via posterior sampling using diffusion priors (Section~\ref{sec:background-backward}).

\subsection{Linear approximation}
\label{sec:background-forward}

Let $\Omega\subset\mathbb R^3$ be the torso volume conductor between the heart surface $\Gamma_{\mathrm H}$ and the body surface $\Gamma_{\mathrm B}$, with $\partial\Omega=\Gamma_{\mathrm H}\cup\Gamma_{\mathrm B}$. At every physical time $t$, the electric potential $\phi(\mathbf r,t)$, with $\mathbf r\in\Omega$, satisfies a stationary conductivity problem described by a linear second-order elliptic boundary-value problem. Suppressing the time argument $t$, we have
\begin{equation}
\nabla\!\cdot(\sigma_{\mathrm c}\nabla\phi)=0
\quad\text{in }\Omega,
\qquad
\phi=\phi_{\mathrm H}
\quad\text{on }\Gamma_{\mathrm H},
\qquad
\mathbf n^{\mathsf T} \sigma_{\mathrm c}\nabla\phi=0
\quad\text{on }\Gamma_{\mathrm B},
\label{eq:laplace-volume-conductor}
\end{equation}
where $\phi_{\mathrm H}$ denotes the prescribed potential on the heart surface $\Gamma_{\mathrm H}$, $\mathbf n$ is the outward unit normal to $\Omega$ on $\Gamma_{\mathrm B}$, and $\sigma_{\mathrm c}$ is the spatially varying scalar electrical conductivity.

For test functions $v$ that vanish on $\Gamma_{\mathrm H}$, integration by parts and the zero-flux condition on $\Gamma_{\mathrm B}$ give the weak form
\begin{equation}\fullskips
\int_{\Omega}
(\nabla v)^{\mathsf T}
\sigma_{\mathrm c}\nabla\phi
\,\mathrm d\Omega
=
0.
\label{eq:laplace-weak-form}
\end{equation}
We approximate $\Omega$ by a tetrahedral mesh and associate each node $j$ with a $P_1$ basis function $\psi_j$. The resulting global stiffness matrix $\mathbf K$, with entries $K_{ij}=\int_{\Omega}(\nabla\psi_i)^{\mathsf T}\sigma_{\mathrm c}\nabla\psi_j\,\mathrm d\Omega$, provides a discrete representation of the conductivity operator~\citep{Wang2011}.

Partitioning $\mathbf K$ according to the unknown free nodes $F$ (in $\Omega$ and on $\Gamma_{\mathrm B}$) and the prescribed heart nodes $H$ (on $\Gamma_{\mathrm H}$) yields the free--free and free--heart blocks $\mathbf K_{\mathrm{FF}}$ and $\mathbf K_{\mathrm{FH}}$. With slight abuse of notation, we reuse $\phi_{\mathrm H}$ to denote the vector of heart-node potentials and write $\phi_{\mathrm F}$ for the vector of free-node potentials. The propagation from the heart nodes to the free nodes at a fixed point in time is then given by
\begin{equation}\fullskips
\mathbf A_\bullet^* = -\mathbf K_{\mathrm{FF}}^{-1}
\mathbf K_{\mathrm{FH}},\qquad
\phi_{\mathrm F}
= \mathbf A_\bullet^*\phi_{\mathrm H},
\label{eq:laplace-weak-form-solution}
\end{equation}
which we aggregate across $T$ time points using $\Phi_{\mathrm F}=[\phi_{\mathrm F,1},\ldots,\phi_{\mathrm F,T}]$ and $\Phi_{\mathrm H}=[\phi_{\mathrm H,1},\ldots,\phi_{\mathrm H,T}]$. Letting $\Phi_{\mathrm B}$ denote the restriction of $\Phi_{\mathrm F}$ to the potentials on the body surface $\Gamma_{\mathrm{B}}$ and $\mathbf A^*$ the corresponding row restriction of $\mathbf A_\bullet^*$, we obtain the linear observation model
\begin{equation}
\Phi_{\mathrm B}  = \mathbf A^*\Phi_{\mathrm H} + \boldsymbol\varepsilon,
\label{eq:generic-forward-operator}
\end{equation}
where $\boldsymbol\varepsilon$ is additive measurement noise. The inverse problem is ill-posed because the volume conductor attenuates and spatially smooths the cardiac field, so inversion amplifies measurement noise and model errors in weakly informed directions, requiring prior information or regularization for stable recovery~\citep{MacLeodBrooks1998}.

\subsection{Posterior sampling-based inversion}\label{sec:background-backward}

Posterior sampling-based inversion combines a learned generative prior $p_{\theta}(\Phi_{\mathrm{H}})$ with a known observation model $\Phi_{\mathrm{B}}=\mathcal{F}(\Phi_{\mathrm{H}})+\boldsymbol\varepsilon$ through Bayes’ rule, $p(\Phi_{\mathrm{H}}\,|\,\Phi_{\mathrm{B}})\propto p(\Phi_{\mathrm{B}}\,|\,\Phi_{\mathrm{H}})p_\theta(\Phi_{\mathrm{H}})$, and reconstructs $\Phi_{\mathrm{H}}$ by drawing samples from $p(\Phi_{\mathrm{H}}\mid \Phi_{\mathrm{B}})$. We recap a common approach with a diffusion prior and a spectral sampler for the linear model $\mathcal{F}(\Phi_{\mathrm{H}})=\mathbf A^*\Phi_{\mathrm{H}}$.

{\bf Diffusion prior~\citep{SalimansHo2022}.} Starting at $\Phi_{\mathrm{H}}^{(0)}=\Phi_{\mathrm{H}}$, a variance-preserving diffusion prior with noise schedule $\{\bar\alpha^{(\tau)}\}_{\tau=1}^S$ models the reverse of the forward process  
\begin{equation}
\Phi_{\mathrm{H}}^{(\tau)}
=
\sqrt{\bar\alpha^{(\tau)}}\,\Phi_{\mathrm{H}}^{(0)}
+
\sqrt{1-\bar\alpha^{(\tau)}}\,\boldsymbol\xi,
\qquad
\xi_{i,t}\overset{\mathrm{i.i.d.}}{\sim}\mathcal N(0,1)
\label{eq:forward-diffusion}
\end{equation}
by predicting the velocity 
\begin{equation}
\mathbf v^{(\tau)}
=
\sqrt{\bar\alpha^{(\tau)}}\,\boldsymbol\xi
-
\sqrt{1-\bar\alpha^{(\tau)}}\,\Phi_{\mathrm{H}}^{(0)}
\end{equation}
with a neural network $\mathbf v_\theta(\Phi_{\mathrm{H}}^{(\tau)},\tau)$, which yields the denoised estimate
\begin{equation}
\widehat{\Phi}_{\mathrm{H}}^{(0)}(\Phi_{\mathrm{H}}^{(\tau)})
=
\sqrt{\bar\alpha^{(\tau)}}\,\Phi_{\mathrm{H}}^{(\tau)}
-
\sqrt{1-\bar\alpha^{(\tau)}}\,\mathbf v_\theta(\Phi_{\mathrm{H}}^{(\tau)},\tau).
\label{eq:denoised-estimate}
\end{equation}
Unconditional sampling starts from $\Phi_{\mathrm{H}}^{(S)}\sim\mathcal N(\mathbf 0,\mathbf I)$ and traverses a decreasing sequence of noise levels using DDIM updates~\citep{Song2021DDIM}, ultimately yielding a draw from the prior $p_\theta(\Phi_{\mathrm{H}})$.

{\bf Spectral sampler~\citep{Kawar2022}.} Denoising Diffusion Restoration Models (DDRM) turn this prior into a solver for linear inverse problems by combining it with an observation model of the form in Equation~\eqref{eq:generic-forward-operator} that defines a likelihood. Assuming i.i.d. Gaussian noise with variance $\sigma^2$, DDRM works in the singular coordinates of
$\mathbf A^*=\mathbf U\boldsymbol\Sigma\mathbf V^{\mathsf T}$. For each singular value $s_i>0$, let $\mathbf u_i$
and $\mathbf v_i$ denote the corresponding left and right singular vectors.
Because the same spatial operator $\mathbf A^*$ acts on every time point, its singular-coordinate transformation applies to each column of the sequence. For each singular direction $i$, we define
\begin{equation}
\bar{\Phi}_{\mathrm{H},i} = 
\mathbf v_i^{\mathsf T} \Phi_{\mathrm{H}},
\qquad
\bar{\Phi}_{\mathrm{B},i}=
\frac{\mathbf u_i^{\mathsf T}\Phi_{\mathrm{B}}}{s_i},
\end{equation}
which yields
\begin{equation}
\bar{\Phi}_{\mathrm{B},i}
=
\bar{\Phi}_{\mathrm{H},i}
+
\frac{\sigma}{s_i}\,\boldsymbol\zeta_{i},
\qquad
\boldsymbol\zeta_{i}\sim\mathcal N(\mathbf 0,\mathbf I_T),
\label{eq:ddrm-spectral}
\end{equation}
i.e., the standard deviation of the effective observation noise in singular direction $i$ is $\sigma/s_i$. At every diffusion step, the DDRM reverse update blends the prior prediction $\widehat{\Phi}_{\mathrm{H}}^{(0)}$ from \eqref{eq:denoised-estimate} with $\bar{\Phi}_{\mathrm{B},i}$ according to whether the current diffusion noise exceeds this scale. If it does, the update moves towards $\bar{\Phi}_{\mathrm{B},i}$; otherwise, the prior prediction dominates. Directions with small singular values therefore receive weaker measurement guidance, directions in the null space of $\mathbf A^*$ are determined by the prior alone, and for $\sigma=0$ the sampler enforces exact consistency with the supplied operator $\mathbf A^*$ in every observed coordinate.

\section{Posterior sampling under operator discrepancy}\label{sec:method}

\begin{figure}[t]
    \centering
    \begin{subfigure}[b]{0.49\textwidth}
        \centering
        \includegraphics[width=\linewidth]{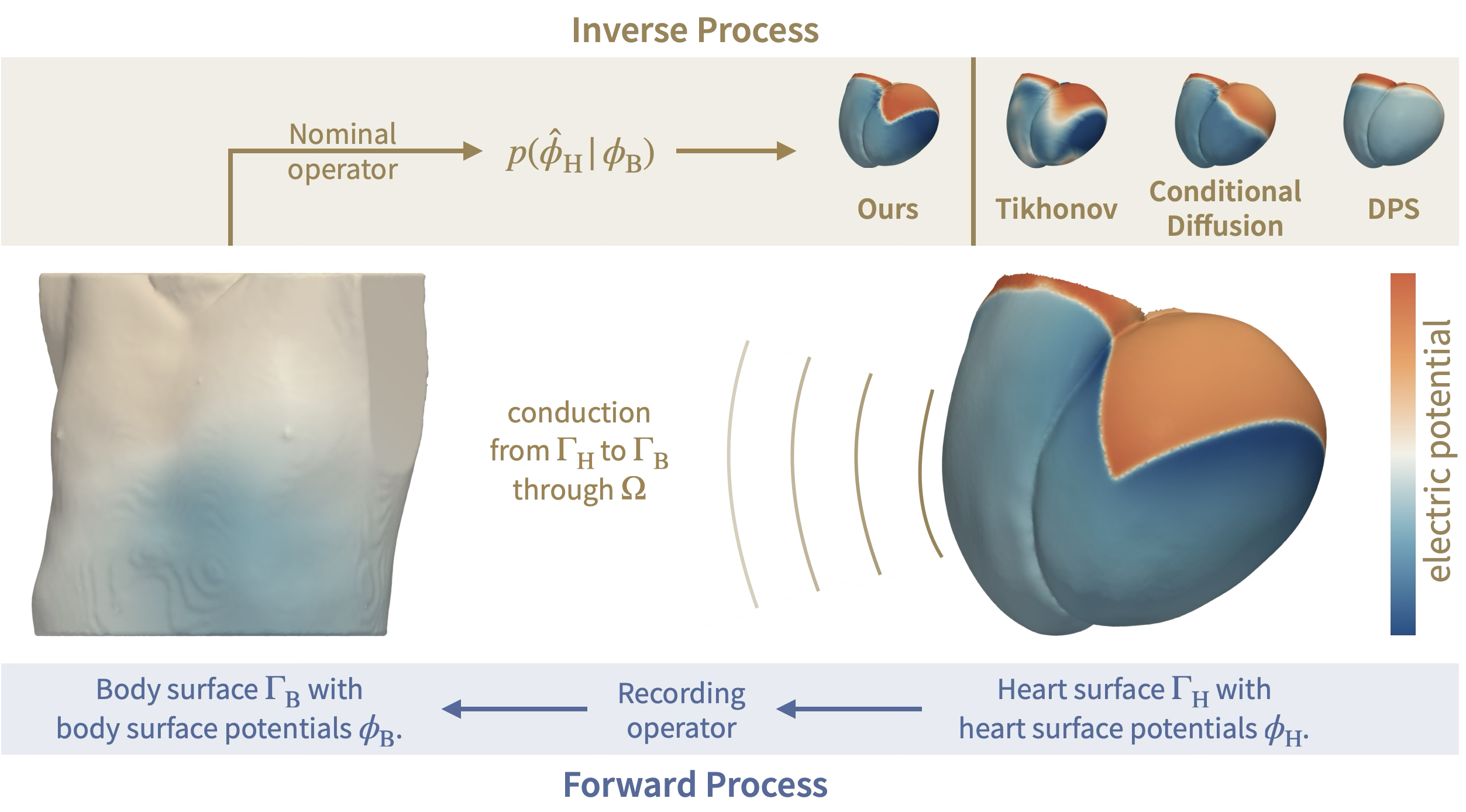}
        \caption{Inversion under operator discrepancy.}
        \label{fig:teaser_task}
    \end{subfigure}%
    \hfill
    \begin{subfigure}[b]{0.49\textwidth}
        \centering
        \raisebox{0.2cm}{\includegraphics[width=\linewidth]{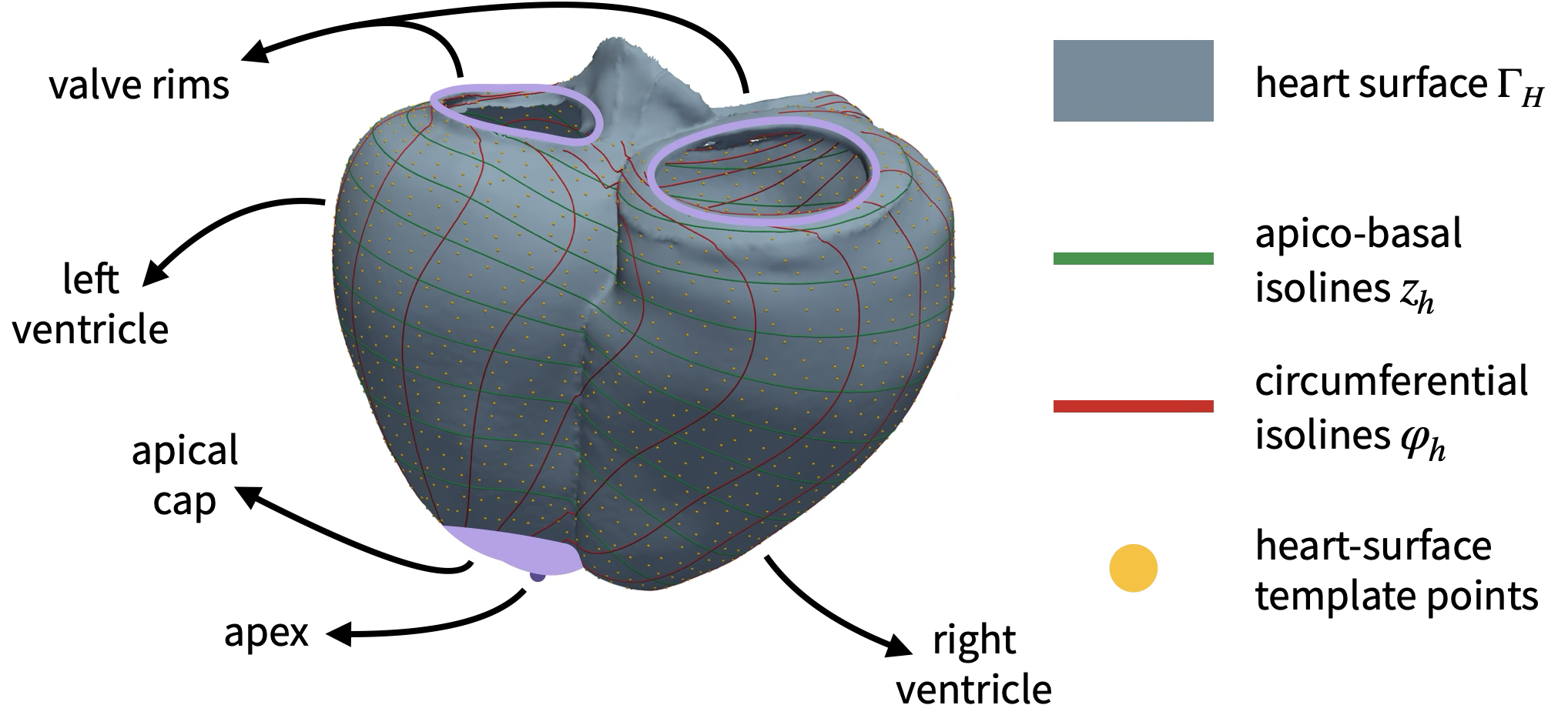}}
        \caption{Heart-surface coordinates.}
        \label{fig:teaser_coordinates}
    \end{subfigure}
    \caption{\textbf{Overview}. 
(\textbf{a}) We consider the task of recovering latent heart-surface potentials $\phi_{\mathrm H}$ from observed body-surface potentials $\phi_{\mathrm B}$. Due to environmental variability, the true (``recording'') operator describing the forward process might differ from the theoretical (``nominal'') operator employed in the inverse process, giving rise to a signal-dependent \emph{operator discrepancy}. (\textbf{b}) For modeling purposes, we represent the heart-surface potentials $\phi_{\mathrm H}$ at $1800$ discrete template points, shown in orange. Each template point is parameterized by an apico-basal coordinate $z_h$ and a circumferential coordinate $\varphi_h$.}
    \label{fig:teaser}
\end{figure}

We build upon the formulations in Section~\ref{sec:background} by introducing geometry-invariant template representations (Section~\ref{sec:templates}). We then present the central concept of operator discrepancy (Section~\ref{sec:forward-operators}) and demonstrate its integration into the likelihood while keeping the prior fixed (Section~\ref{sec:approximation-error-likelihood}).     
\subsection{Template representations}
\label{sec:templates}

Since the node locations associated with the potentials $\Phi_{\mathrm H}$ and $\Phi_{\mathrm B}$ depend on the heart and body geometry, they are subject-specific. To account for inter-subject variation, we introduce shared sets of template points on $\Gamma_{\mathrm H}$ and $\Gamma_{\mathrm B}$. We denote the potential values at the $n_{\mathrm H}$ heart-template points by $\mathbf h\in\mathbb R^{n_{\mathrm H}\times T}$ and the observed potential values at the $n_{\mathrm B}$ body-template points by $\mathbf b\in\mathbb R^{n_{\mathrm B}\times T}$, with one time point per column. The template points reference equivalent locations across subjects, with two subject-specific linear maps connecting the shared templates to the native meshes: the heart-template map $\mathbf T_{\mathrm H}$ yields $\Phi_{\mathrm H}$ from $\mathbf h$ and the body-template map $\mathbf T_{\mathrm B}$ maps $\Phi_{\mathrm B}$ to $\mathbf b$. Extending~\eqref{eq:generic-forward-operator}, our observation model is thus 
\begin{equation}
\mathbf A = \mathbf T_{\mathrm B} \mathbf A^* \mathbf T_{\mathrm H}, \quad \mathbf b = \mathbf A\mathbf h+\boldsymbol\varepsilon.
\label{eq:specific-forward-operator}
\end{equation}

{\bf Heart template.} We describe each point on the heart surface $\Gamma_{\mathrm H}$ using an apico-basal coordinate $z_h$ and a periodic circumferential coordinate $\varphi_h$, such that equal-sized regions in the $(z_h,\varphi_h)$ domain correspond to approximately equal heart-surface areas. The $n_{\mathrm H}=1800$ template points form a regular $30\times 60$ grid in this coordinate system, which we construct on one reference heart and then transfer to other subjects through known node correspondences. 

The prescribed heart-mesh potentials $\Phi_{\mathrm H}$ are mapped to the heart-template potentials $\mathbf h\in\mathbb R^{1800\times T}$ by a mass-weighted $L^2$ fit over the represented surface regions, applied independently at each time point. Conversely, for the forward problem, the subject-specific linear map $\mathbf T_{\mathrm H}$ yields the prescribed heart-mesh potentials $\Phi_{\mathrm H}$ by interpolating the heart-template potentials $\mathbf h$ at the heart-mesh nodes. Both the template and its subject-specific maps remain fixed across operator conditions. Figure~\ref{fig:teaser_coordinates} illustrates the construction, with mathematical details provided in Appendix~\ref{sec:chart-construction}.

{\bf Body template.} The body-template potentials $\mathbf b$ are observed at $n_{\mathrm B}$ electrode locations on the body surface $\Gamma_{\mathrm B}$. Since potentials are only defined up to an additive constant, we introduce the centering matrix $\mathbf C=\mathbf I-\frac{\mathbf1\mathbf1^{\mathsf T}}{n_{\mathrm B}}$ and define the body-template map as $\mathbf T_{\mathrm B}=\mathbf C\mathbf S$, where $\mathbf S$ selects the body-mesh potentials $\Phi_{\mathrm B}$ at the electrode locations. Consistent with this centering, we draw $\mathbf z_t\sim\mathcal N(\mathbf 0,\sigma^2\mathbf I_{n_{\mathrm B}})$ independently for each time point and set $\boldsymbol\varepsilon_t=\mathbf C\mathbf z_t$, so $\operatorname{Cov}(\boldsymbol\varepsilon_t)=\sigma^2\mathbf C$.

The covariance $\sigma^2\mathbf C$ is rank-deficient, because $\mathbf C$ removes one degree of freedom, and this violates the i.i.d.\ Gaussian noise assumption of the spectral posterior sampler. The centered noise instead lies in the $(n_{\mathrm B}{-}1)$-dimensional subspace of zero-mean measurements. We express $\mathbf b$, $\mathbf A$, and $\boldsymbol\varepsilon$ in an orthonormal basis of this subspace and, with slight abuse of notation, reuse the same symbols for these reduced counterparts, so that the reduced noise becomes isotropic, $\boldsymbol\varepsilon_t\sim\mathcal N(\mathbf 0,\sigma^2\mathbf I_{n_{\mathrm B}-1})$. Appendix~\ref{sec:coordinate-reduction} gives the details.

\subsection{Operator discrepancy}
\label{sec:forward-operators}
Section~\ref{sec:background-forward} defines a generic forward operator $\mathbf A^*$ for a given torso geometry and conductivity distribution. In practice, the operator available at reconstruction time is a \emph{nominal} operator $\mathbf A$, whereas the operator producing the observation at measurement time is a potentially different \emph{recording} operator $\mathbf A_0$. When $\mathbf A_0\neq\mathbf A$, the reconstruction method consumes an observation produced by $\mathbf A_0$ while only knowing $\mathbf A$. Accordingly, the nominal observation model in Equation~\eqref{eq:specific-forward-operator} is formulated in terms of $\mathbf A$, despite the observation being generated by $\mathbf A_0$. OpDis-PS accounts for the resulting mismatch inside the likelihood. 

The reduced observation generated by the recording operator $\mathbf A_0$ can be written in terms of the nominal operator $\mathbf A$ available during reconstruction as
\begin{equation}
\mathbf b
=
\mathbf A_0\mathbf h+\boldsymbol\varepsilon
=
\mathbf A\mathbf h
+
\underbrace{(\mathbf A_0-\mathbf A)\mathbf h}_{\mathbf d}
+
\boldsymbol\varepsilon,
\qquad
\boldsymbol\varepsilon_t
\sim
\mathcal N(\mathbf 0,\sigma^2\mathbf I_{n_{\mathrm B}-1}).
\label{eq:operator-perturbation}
\end{equation}
The coordinate reduction gives i.i.d.\ Gaussian measurement noise, as assumed by the spectral posterior sampler. The discrepancy $\mathbf d$ vanishes in the nominal condition.

\subsection{Approximation-error likelihood}
\label{sec:approximation-error-likelihood}

\begin{figure}[t]
    \centering
    \includegraphics[width=1\linewidth]{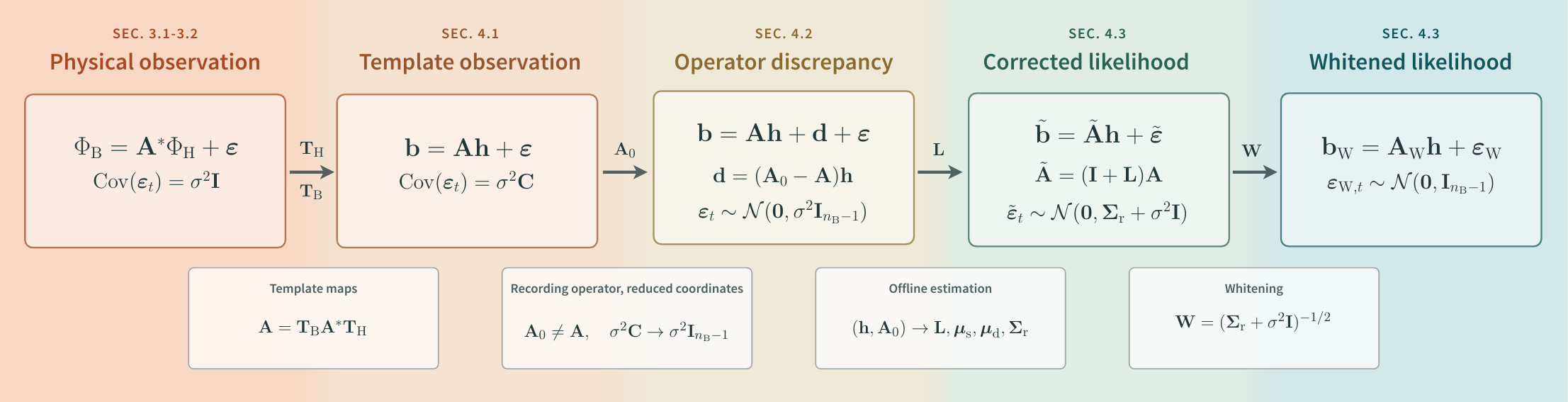}
    \caption{{\bf Model}. The diagram traces the observation model from the physical forward problem to the whitened likelihood consumed by the posterior sampler. The recording operator $\mathbf A_0$ generates the observation, while the nominal operator $\mathbf A$ is used for reconstruction; from the discrepancy panel onward, all quantities are expressed in reduced coordinates.}
    \label{fig:architecture}
\end{figure}

We model the reduced operator discrepancy $\mathbf d$ of Equation~\eqref{eq:operator-perturbation} with a signal-conditional variant of the Bayesian approximation-error framework \citep{KaipioSomersalo2007}. Specifically, we approximate its dependence on the heart-template potentials $\mathbf h$ through the nominal signal $\mathbf s=\mathbf A\mathbf h$.

For each physical time point $t$, we model the corresponding columns $\mathbf s_t$ and $\mathbf d_t$ as jointly Gaussian,
\begin{equation}
\begin{pmatrix}
\mathbf s_t\\
\mathbf d_t
\end{pmatrix}
\sim
\mathcal N\!\left(
\begin{pmatrix}
\boldsymbol\mu_{\mathrm s}\\
\boldsymbol\mu_{\mathrm d}
\end{pmatrix},
\begin{pmatrix}
\mathbf C_{\mathrm{ss}} & \mathbf C_{\mathrm{sd}}\\
\mathbf C_{\mathrm{ds}} & \mathbf C_{\mathrm{dd}}
\end{pmatrix}
\right),
\label{eq:joint-gaussian}
\end{equation}
with the same means and covariance blocks across time. The joint model implies
\begin{equation}
\mathbf d_t\mid\mathbf s_t
\sim
\mathcal N\!\left(
\boldsymbol\mu_{\mathrm d}+\mathbf L(\mathbf s_t-\boldsymbol\mu_{\mathrm s}),
\boldsymbol\Sigma_{\mathrm r}
\right),
\label{eq:conditional-aem}
\end{equation}
with linear predictor $\mathbf L=\mathbf C_{\mathrm{ds}}\mathbf C_{\mathrm{ss}}^{-1}$ and residual covariance $\boldsymbol\Sigma_{\mathrm r}=\mathbf C_{\mathrm{dd}}-\mathbf L\mathbf C_{\mathrm{sd}}$. Under the joint-Gaussian model, the residual $\mathbf r_t=\mathbf d_t-\boldsymbol\mu_{\mathrm d}-\mathbf L(\mathbf s_t-\boldsymbol\mu_{\mathrm s})$ follows $\mathcal N(\mathbf 0,\boldsymbol\Sigma_{\mathrm r})$ and is independent of the signal $\mathbf s_t$. The residual covariance is shared across time and our approximate likelihood treats the residuals as independent across time; see Appendix~\ref{sec:signal-conditional-derivation} for a full derivation.

{\bf Corrected likelihood.} We estimate the model parameters separately for each discrepancy source, using training samples of $\mathbf h$ paired with recording operators $\mathbf A_0$ drawn from the corresponding perturbation distribution. In practice, we compute $\mathbf L$ using trace-scaled ridge regularization of $\mathbf C_{\mathrm{ss}}$ and obtain $\boldsymbol\Sigma_{\mathrm r}$ from the fitted residuals using Ledoit--Wolf shrinkage \citep{LedoitWolf2004}; details are provided in
Appendix~\ref{sec:opdis-discrepancy-estimation}.

We absorb the conditional-mean predictor into a corrected operator $\tilde{\mathbf A}=(\mathbf I+\mathbf L)\mathbf A$ and subtract the constant bias $\boldsymbol\mu_{\mathrm d}-\mathbf L\boldsymbol\mu_{\mathrm s}$ from each column of $\mathbf b$ to obtain the bias-corrected observation $\tilde{\mathbf b}$. The corrected observation model reads $\tilde{\mathbf b}=\tilde{\mathbf A}\mathbf h+\tilde{\boldsymbol\varepsilon}$, where the combined noise has columns $\tilde{\boldsymbol\varepsilon}_t=\mathbf r_t+\boldsymbol\varepsilon_t\sim\mathcal N(\mathbf 0,\boldsymbol\Sigma_{\mathrm r}+\sigma^2\mathbf I_{n_{\mathrm B}-1})$. Only the residual mismatch enters the noise distribution.

{\bf Whitened likelihood.} The combined noise is correlated across measurement coordinates. \mbox{Following}~\citep{Nezakati2026CARD}, we whiten the corrected observation and operator to obtain the isotropic Gaussian noise model required by the spectral posterior sampler. With $\mathbf W=(\boldsymbol\Sigma_{\mathrm r}+\sigma^2\mathbf I_{n_{\mathrm B}-1})^{-1/2}$, we set $\mathbf A_{\mathrm W}=\mathbf W\tilde{\mathbf A}$ and $\mathbf b_{\mathrm W}=\mathbf W\tilde{\mathbf b}$. The whitened noise $\boldsymbol\varepsilon_{\mathrm W}=\mathbf W\tilde{\boldsymbol\varepsilon}$ has columns $\boldsymbol\varepsilon_{\mathrm{W},t}\sim\mathcal N(\mathbf 0,\mathbf I_{n_{\mathrm B}-1})$, and the whitened observation model reads
\begin{equation}
\mathbf b_{\mathrm W}
=
\mathbf A_{\mathrm W}\mathbf h+\boldsymbol\varepsilon_{\mathrm W}.
\label{eq:conditional-whitening}
\end{equation}

We combine the whitened likelihood with a frozen unconditional prior trained on heart-template potentials. The prior remains fixed across discrepancy conditions. Appendix~\ref{sec:unconditional-prior-architecture} describes its architecture. For each observation, the spectral posterior sampler consumes $(\mathbf A_{\mathrm W},\mathbf b_{\mathrm W})$ and returns $K$ approximate posterior draws. Their Monte Carlo mean defines the point estimate. Figure~\ref{fig:architecture} illustrates the resulting pipeline.

\section{Experiments}
\label{sec:experiments}

We first describe the datasets and operator-discrepancy settings (Section~\ref{sec:data}). We then introduce the evaluated baselines (Section~\ref{sec:baselines}) and define the metrics used for quantitative comparison (Section~\ref{sec:metrics}). Finally, we report quantitative results (Section~\ref{sec:quan_results}) and qualitative insights (Section~\ref{sec:qual_results}).

\subsection{Data}
\label{sec:data}

{\bf Geometries and physiological simulations.}
We use 1000 synthetic four-chamber heart meshes from a CT-based statistical shape model \citep{Rodero2021} and retain the left and right ventricles. Their outer surface defines the heart surface used for reconstruction. Torso anatomies originate from 1228 CT scans in the TotalSegmentator dataset \citep{Wasserthal2023Dataset,Wasserthal2023}. We retain 167 torso geometries with complete coverage, an extracted body surface, and the lungs included in the volume conductor. We combine each heart with multiple anatomically compatible torso geometries and register it to the ventricular envelope extracted from the corresponding torso CT. A fixed vest layout provides $n_{\mathrm B}=176$ body-surface electrode locations. We split heart and torso identities before pairing, resulting in 1700 training pairs, 100 validation pairs, and 250 physiological test pairs.

We simulate ventricular electrophysiology with openCARP \citep{Plank2021openCARP} following the MedalCare-XL protocol \citep{Gillette2023MedalCare}, which couples a pseudo-bidomain formulation with Mitchell--Schaeffer ionic dynamics, transversely isotropic conduction along the myocardial fibers, and a fast endocardial layer (Appendix~\ref{sec:simulation-details}); each simulation spans $450$~ms sampled every $5$~ms. Three physiological rounds resample the action-potential-duration field and five endocardial initiation sites, producing 5100 training sequences over the QRST interval, which covers ventricular depolarization and repolarization. We map the simulated potentials to the heart-template representation $\mathbf h\in\mathbb R^{1800\times91}$ defined in Section~\ref{sec:templates}. We generate the noise-free training and validation body-template potentials directly as $\mathbf A_0\mathbf h$, with $\mathbf A_0=\mathbf A$ in these nominal splits.

{\bf Operator discrepancy.}
We evaluate the same 250 physiological test pairs in one matched-operator reference and two operator-discrepancy test sets. In the matched-operator reference, the recording and reconstruction operators coincide, $\mathbf A_0=\mathbf A$. In each discrepancy test set, the recording operator $\mathbf A_0$ produces the observations, while every method with an explicit forward operator receives only the nominal operator $\mathbf A$. The heart--torso pairs and electrode locations remain fixed across conditions.

For lung-conductivity discrepancy, we model variability associated with different lung inflation states. The nominal operator $\mathbf A$ assigns piecewise-constant conductivities to two tissue compartments, the lungs ($0.05$~S/m) and the remaining torso ($0.22$~S/m). For each test case, the recording operator $\mathbf A_0$ uses a lung conductivity drawn from $\mathcal U[0.03,0.15]$~S/m while keeping the torso conductivity fixed \citep{Nopp1993,Keller2010}. The approximation-error model uses operators with lung conductivities drawn from the narrower interval $[0.04,0.10]$~S/m.

For subcutaneous-fat discrepancy, the nominal operator $\mathbf A$ treats the torso outside the lungs as homogeneous, whereas the recording operator $\mathbf A_0$ includes a subcutaneous layer with conductivity $0.04$~S/m directly beneath the body surface. Its thickness depends on the torso-size category and ranges from $3$ to $20$~mm across the test cases. The approximation-error model uses operators with thicknesses drawn uniformly from $[5,15]$~mm.

\begin{figure}[t]
    \centering
    \begin{subfigure}[b]{0.27\textwidth}
        \vspace{0pt}
        \centering
        \includegraphics[width=0.9\linewidth]{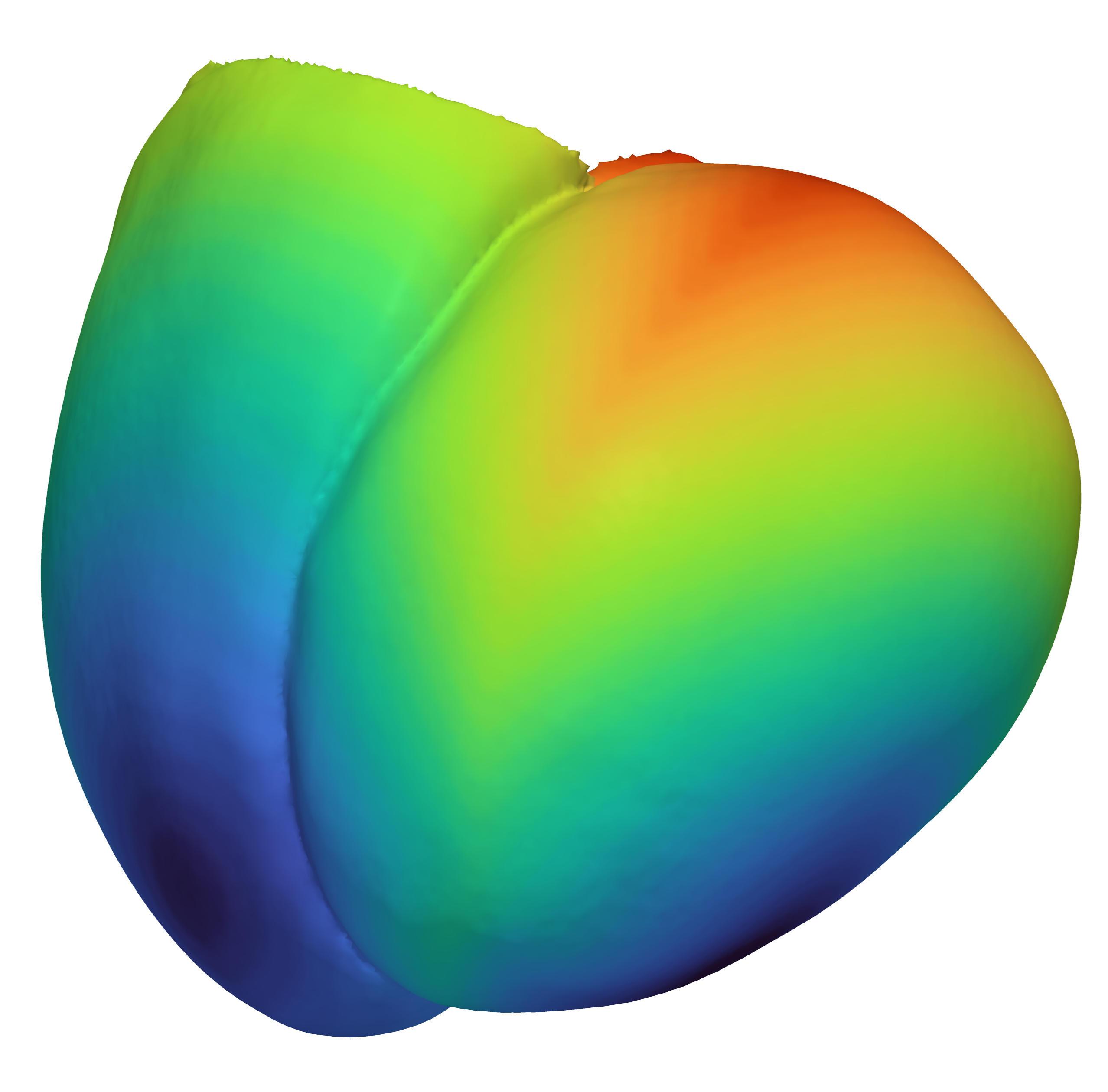}
        \caption{Left bundle branch block.}
        \label{fig:a}
    \end{subfigure}
    \qquad
    \begin{subfigure}[b]{0.27\textwidth}
        \vspace{0pt}
        \centering
        \includegraphics[width=0.9\linewidth]{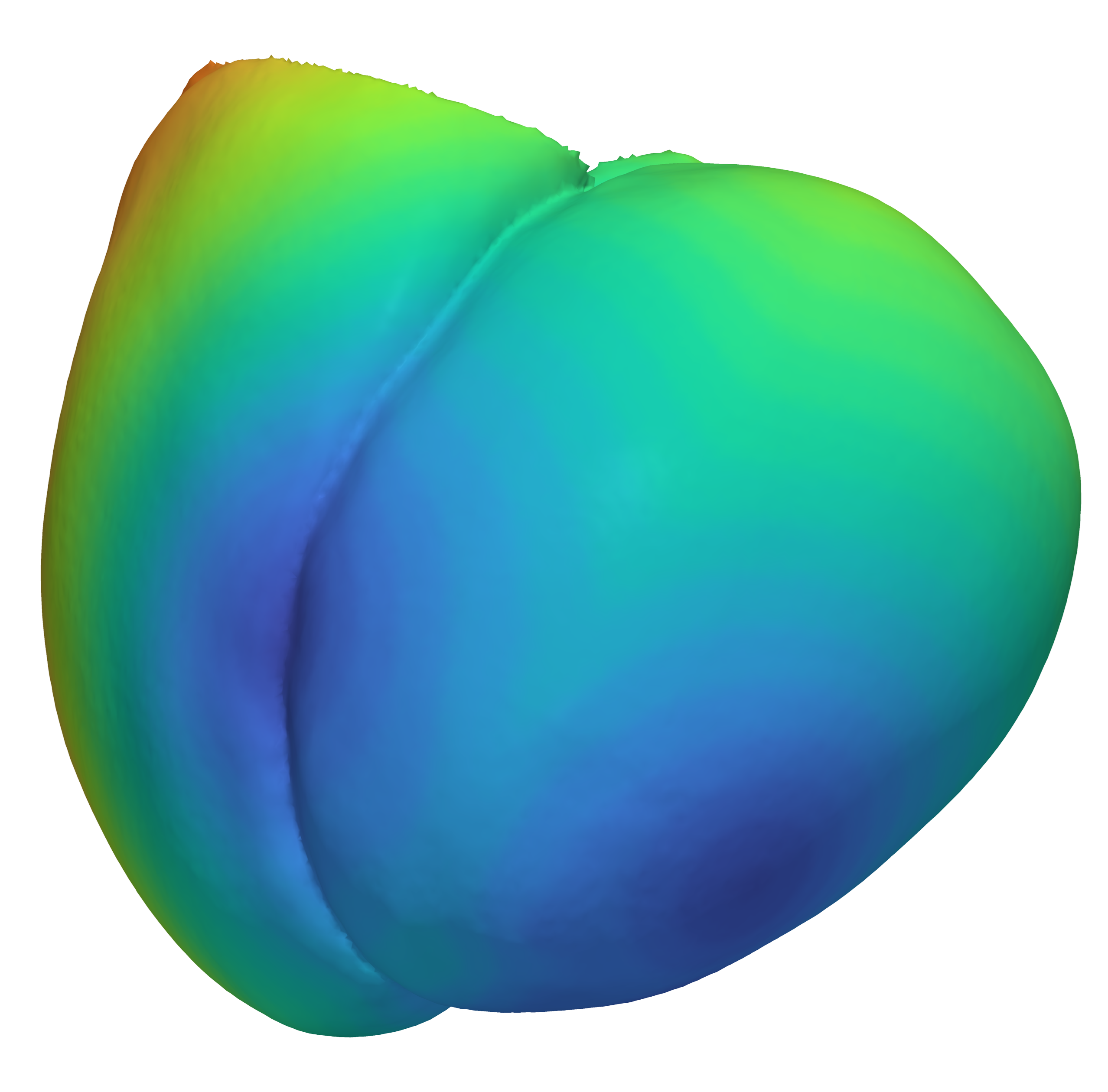}
        \caption{Right bundle branch block.}
        \label{fig:b}
    \end{subfigure}
    \qquad
    \begin{subfigure}[b]{0.27\textwidth}
        \vspace{0pt}
        \centering
        \raisebox{0.1cm}{\includegraphics[width=0.838\linewidth]{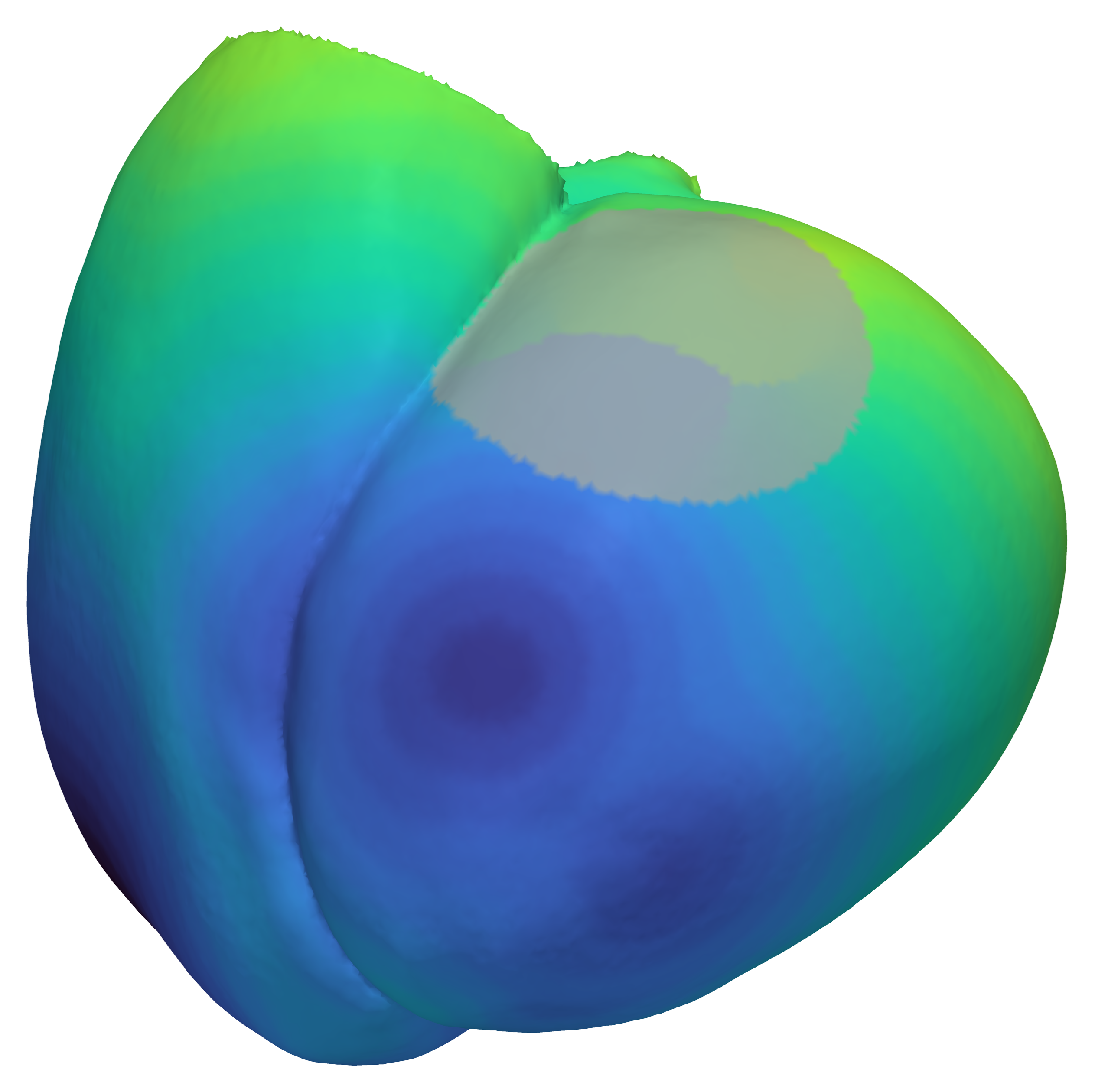}}
        \caption{Myocardial scar.}
        \label{fig:c}
    \end{subfigure}
    \hfill
    \begin{minipage}[b]{0.03\textwidth}
        \vspace{0.85cm}
        \centering
        \raisebox{0.74cm}{\includegraphics[height=2.96cm,width=0.7\linewidth]{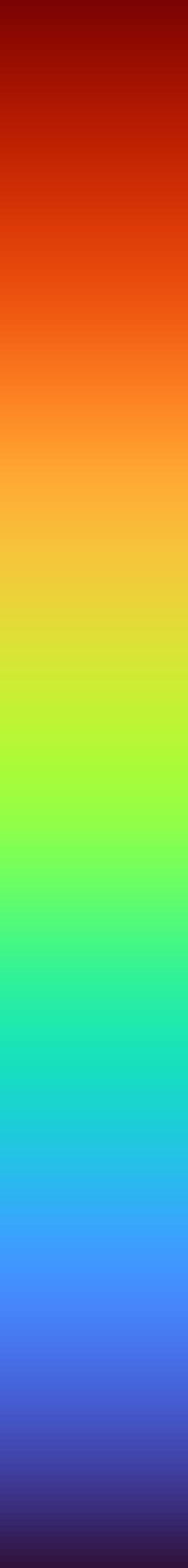}}
    \end{minipage}

    \caption{{\bf Illustrative examples of pathologies.} Ground-truth local activation-time (LAT) maps for the selected pathologies. The gray overlay marks the modeled scar region. The bundle-branch-block examples share the same heart geometry, whereas the scar example has a different anatomy.}
    \label{fig:pathologies}
\end{figure}

{\bf Pathological data.}
In addition to physiological data, we evaluate four severity configurations for each of left bundle branch block (LBBB), right bundle branch block (RBBB), and myocardial scar. We simulate every configuration on 60 heart--torso geometries from the test split, yielding 720 pathological sequences per operator-discrepancy condition. No pathological sequence enters training or validation. Figure~\ref{fig:pathologies} illustrates the three pathology types, and Table~\ref{tab:pathology-configurations} lists all configurations.

{\bf Measurement noise.}
We draw Gaussian noise independently over electrodes and time and apply the centering matrix $\mathbf C$ from Section~\ref{sec:templates}. We choose its standard deviation to target the sequence-level Frobenius signal-to-noise ratio (SNR)
\begin{equation}
\operatorname{SNR}_{\mathrm{dB}}
=
20\log_{10}
\frac{\lVert\mathbf A_0\mathbf h\rVert_F}
     {\lVert\boldsymbol\varepsilon\rVert_F}.
\end{equation}
We use 40 dB for the main comparison to emphasize operator discrepancy and 30 dB for an additional evaluation under stronger measurement noise. Appendix~\ref{sec:discrepancy-magnitude} quantifies the relative magnitudes of measurement noise and operator discrepancy underlying this choice.

\subsection{Baselines} \label{sec:baselines}

We consider baselines from three categories: variational regularization, amortized inference, and posterior sampling. All baselines are evaluated on the same test cases as OpDis-PS, and every method with an explicit forward operator uses the nominal operator $\mathbf A$. Appendix~\ref{sec:architectural-details} details the architectures of the data-driven methods and Appendix~\ref{sec:hyperparameters} reports the full settings for all methods.

{\bf Variational regularization.}
First-order generalized Tikhonov penalizes the template surface gradient and selects $\lambda$ case-wise by generalized cross-validation (GCV) \citep{MessingerRapportRudy1988MathBiosci,Golub1979}. Spatio-temporal Tikhonov adds a second-order temporal-difference penalty \citep{Brooks1999}. %
Both variants discard reduced-Gram eigenmodes with $w_i\leq10^{-6}w_{\max}$, making them truncated singular value decomposition (TSVD)--Tikhonov hybrids. Our Total Variation baseline applies the spatial template-gradient penalty \citep{RudinOsherFatemi1992} to each time point independently and uses the Chambolle--Pock primal--dual algorithm \citep{ChambollePock2011}. %

{\bf Amortized inference.}
We consider a transformer and a conditional diffusion model. Neither baseline explicitly uses the forward operator during inference. Both are trained on nominal pairs with average-referenced Gaussian noise augmentation and use nominal validation pairs for checkpoint selection. 
Both models map body-template potentials to heart-template potentials. To this end, the transformer baseline implements the architecture of~\citet{Vaswani2017}, whereas the diffusion baseline uses a Palette-style conditional diffusion model~\citep{Saharia2022Palette}, following recent geometry-free conditional diffusion in ECGI~\citep{JaraMeyers2026}. %

{\bf Posterior sampling.} We consider DDRM~\citep{Kawar2022} and DPS~\citep{Chung2023}. DDRM combines a frozen unconditional prior over heart-template potentials based on denoising diffusion probabilistic models~\citep{Ho2020} with the nominal operator $\mathbf A$ and an additive noise model. DPS uses the same prior and nominal operator $\mathbf A$ for gradient-based guidance, scaled by a weight selected on the validation set.

\subsection{Metrics}
\label{sec:metrics}

We evaluate reconstructed heart-template potentials $\widehat{\mathbf h}$ against the ground truth $\mathbf h$ using five complementary metrics. The relative error (RE) measures global reconstruction error, while the spatial and temporal correlation coefficients (sCC, tCC) measure spatial-pattern and temporal-waveform agreement. The activation time error (ATE) quantifies local activation-time differences in milliseconds, and the localization error (LE) gives the distance between the reconstructed and true earliest activation sites on the heart surface in millimeters. Formal definitions and the statistical testing protocol are provided in Appendix~\ref{sec:metric-definitions}.
\subsection{Quantitative results}
\label{sec:quan_results}

\begin{table}[t]
\centering
\caption{{\bf Quantitative results (pathological data; 40 dB).}
We evaluate two sources of operator discrepancy (lung conductivity and subcutaneous fat) on 720 pathological test cases at 40 dB against variational regularization (R), amortized inference (A), and posterior sampling (P) baselines. Data-driven results are mean $\pm$ standard deviation over five training runs; $\star$ denotes $p<0.05$ for the best versus second-best data-driven method (Appendix~\ref{sec:metric-definitions}).}
\label{tab:phy_pat}
\resizebox{\textwidth}{!}{%
\begin{tabular}{cclccccc}
\toprule
\multirow{2}{*}{Discrepancy} & & \multirow{2}{*}{Method} & \multicolumn{5}{c}{Metric} \\
\cmidrule(lr){4-8}
& & & RE\,$\downarrow$ & ATE [ms]\,$\downarrow$ & LE [mm]\,$\downarrow$ & sCC\,$\uparrow$ & tCC\,$\uparrow$ \\
\midrule
\multirow{8}{*}{\rotatebox[origin=c]{90}{\shortstack{lung\\conductivity}}}
& \multirow{3}{*}{R}
& Tikhonov (spatial)~\citep{MessingerRapportRudy1988MathBiosci}
    & $0.737$
    & $12.058$
    & $23.878$
    & $0.646$
    & $0.742$ \\
&
& Tikhonov (spatio-temporal)~\citep{Brooks1999}
    & $0.720$
    & $7.927$
    & $9.786$
    & $0.667$
    & $0.755$ \\
&
& Total Variation~\citep{RudinOsherFatemi1992}
    & $0.618$
    & $8.559$
    & $11.470$
    & $0.733$
    & $0.763$ \\
\cmidrule(lr){2-8}
&
\multirow{2}{*}{A}
& Transformer~\citep{Vaswani2017}
    & $0.561 \pm 0.002$
    & $7.143 \pm 0.080$
    & $7.349 \pm 0.251$
    & $0.746 \pm 0.005$
    & $0.792 \pm 0.002$ \\
&
& Palette~\citep{Saharia2022Palette}
    & $0.517 \pm 0.003$
    & $6.384 \pm 0.219$
    & $7.940 \pm 0.513$
    & $0.786 \pm 0.002$
    & $0.822 \pm 0.003$ \\
\cmidrule(lr){2-8}
&
\multirow{3}{*}{P}
& DPS~\citep{Chung2023}
    & $0.617 \pm 0.004$
    & $11.214 \pm 0.180$
    & $12.197 \pm 0.566$
    & $0.724 \pm 0.003$
    & $0.739 \pm 0.003$ \\
&
& DDRM~\citep{Kawar2022}
    & $0.425 \pm 0.003$
    & ${\bf 3.599 \pm 0.077}$
    & ${\bf 2.688 \pm 0.070}^{\star}$
    & $0.849 \pm 0.002$
    & $0.896 \pm 0.002$ \\
&
& OpDis-PS ({\bf ours})
    & ${\bf 0.393 \pm 0.003}^{\star}$
    & $3.621 \pm 0.091$
    & $2.927 \pm 0.077$
    & ${\bf 0.853 \pm 0.003}^{\star}$
    & ${\bf 0.903 \pm 0.002}^{\star}$ \\
\midrule
\multirow{8}{*}{\rotatebox[origin=c]{90}{\shortstack{subcutaneous\\fat}}}
& \multirow{3}{*}{R}
& Tikhonov (spatial)~\citep{MessingerRapportRudy1988MathBiosci}
    & $0.784$
    & $11.739$
    & $24.112$
    & $0.630$
    & $0.722$ \\
&
& Tikhonov (spatio-temporal) \citep{Brooks1999}
    & $0.756$
    & $7.640$
    & $9.635$
    & $0.655$
    & $0.738$ \\
&
& Total Variation~\citep{RudinOsherFatemi1992}
    & $0.635$
    & $8.488$
    & $10.439$
    & $0.732$
    & $0.764$ \\
\cmidrule(lr){2-8}
&
\multirow{2}{*}{A}
& Transformer~\citep{Vaswani2017}
    & $0.565 \pm 0.002$
    & $7.106 \pm 0.085$
    & $7.366 \pm 0.268$
    & $0.746 \pm 0.005$
    & $0.792 \pm 0.002$ \\
&
& Palette~\citep{Saharia2022Palette}
    & $0.520 \pm 0.003$
    & $6.495 \pm 0.246$
    & $8.150 \pm 0.662$
    & $0.785 \pm 0.002$
    & $0.820 \pm 0.003$ \\
\cmidrule(lr){2-8}
&
\multirow{3}{*}{P}
& DPS~\citep{Chung2023}
    & $0.616 \pm 0.004$
    & $10.649 \pm 0.169$
    & $10.515 \pm 0.357$
    & $0.729 \pm 0.003$
    & $0.751 \pm 0.003$ \\
&
& DDRM~\citep{Kawar2022}
    & $0.460 \pm 0.003$
    & $3.605 \pm 0.085$
    & $3.383 \pm 0.085$
    & $0.852 \pm 0.002$
    & $0.892 \pm 0.002$ \\
&
& OpDis-PS ({\bf ours})
    & ${\bf 0.395 \pm 0.003}^{\star}$
    & ${\bf 3.602 \pm 0.091}$
    & ${\bf 3.160 \pm 0.086}^{\star}$
    & ${\bf 0.853 \pm 0.003}$
    & ${\bf 0.903 \pm 0.002}^{\star}$ \\
\bottomrule
\multicolumn{8}{l}{\vspace{-0.4cm}\small
RE = relative error; ATE = activation time error; LE = localization error; sCC = spatial corr. coefficient; tCC = temporal corr. coefficient}
\end{tabular}}
\end{table}

\begin{table}[t]
\centering
\caption{{\bf Quantitative results (physiological data; 40 dB).}
We evaluate the same two sources of operator discrepancy (lung conductivity and subcutaneous fat) on 250 physiological test cases at 40 dB against variational regularization (R), amortized inference (A), and posterior sampling (P) baselines. Data-driven results are mean $\pm$ standard deviation over five training runs; $\star$ denotes $p<0.05$ for the best versus second-best data-driven method (Appendix~\ref{sec:metric-definitions}).}
\label{tab:phy_healthy}
\resizebox{\textwidth}{!}{%
\begin{tabular}{cclccccc}
\toprule
\multirow{2}{*}{Discrepancy} & & \multirow{2}{*}{Method} & \multicolumn{5}{c}{Metric} \\
\cmidrule(lr){4-8}
& & & RE\,$\downarrow$ & ATE [ms]\,$\downarrow$ & LE [mm]\,$\downarrow$ & sCC\,$\uparrow$ & tCC\,$\uparrow$ \\
\midrule
\multirow{8}{*}{\rotatebox[origin=c]{90}{\shortstack{lung\\conductivity}}}
& \multirow{3}{*}{R}
& Tikhonov (spatial)
    & $0.765$
    & $7.680$
    & $15.662$
    & $0.587$
    & $0.751$ \\
&
& Tikhonov (spatio-temporal)
    & $0.749$
    & $5.825$
    & $7.408$
    & $0.608$
    & $0.763$ \\
&
& Total Variation
    & $0.583$
    & $6.303$
    & $10.199$
    & $0.745$
    & $0.796$ \\
\cmidrule(lr){2-8}
&
\multirow{2}{*}{A}
& Transformer
    & $0.395 \pm 0.002$
    & $2.822 \pm 0.045$
    & $4.379 \pm 0.192$
    & $0.814 \pm 0.009$
    & $0.905 \pm 0.001$ \\
&
& Palette
    & $0.379 \pm 0.003$
    & $2.738 \pm 0.029$
    & $5.380 \pm 0.338$
    & $0.851 \pm 0.002$
    & $0.910 \pm 0.002$ \\
\cmidrule(lr){2-8}
&
\multirow{3}{*}{P}
& DPS
    & $0.482 \pm 0.004$
    & $4.466 \pm 0.087$
    & $8.643 \pm 0.378$
    & $0.798 \pm 0.006$
    & $0.853 \pm 0.003$ \\
&
& DDRM
    & $0.372 \pm 0.001$
    & $2.221 \pm 0.011$
    & ${\bf 2.108 \pm 0.101}^{\star}$
    & $0.837 \pm 0.004$
    & $0.926 \pm 0.000$ \\
&
& OpDis-PS ({\bf ours})
    & ${\bf 0.313 \pm 0.001}^{\star}$
    & ${\bf 2.033 \pm 0.008}^{\star}$
    & $2.198 \pm 0.146$
    & ${\bf 0.858 \pm 0.003}^{\star}$
    & ${\bf 0.943 \pm 0.000}^{\star}$ \\
\midrule
\multirow{8}{*}{\rotatebox[origin=c]{90}{\shortstack{subcutaneous\\fat}}}
& \multirow{3}{*}{R}
& Tikhonov (spatial)
    & $0.775$
    & $7.351$
    & $18.790$
    & $0.579$
    & $0.741$ \\
&
& Tikhonov (spatio-temporal)
    & $0.750$
    & $5.481$
    & $7.595$
    & $0.606$
    & $0.755$ \\
&
& Total Variation
    & $0.601$
    & $6.320$
    & $9.218$
    & $0.740$
    & $0.795$ \\
\cmidrule(lr){2-8}
&
\multirow{2}{*}{A}
& Transformer
    & $0.399 \pm 0.004$
    & $2.883 \pm 0.056$
    & $4.522 \pm 0.228$
    & $0.810 \pm 0.008$
    & $0.903 \pm 0.002$ \\
&
& Palette
    & $0.376 \pm 0.003$
    & $2.698 \pm 0.025$
    & $4.990 \pm 0.223$
    & $0.854 \pm 0.002$
    & $0.912 \pm 0.002$ \\
\cmidrule(lr){2-8}
&
\multirow{3}{*}{P}
& DPS
    & $0.501 \pm 0.005$
    & $4.513 \pm 0.094$
    & $7.550 \pm 0.380$
    & $0.793 \pm 0.006$
    & $0.851 \pm 0.003$ \\
&
& DDRM
    & $0.409 \pm 0.002$
    & $2.365 \pm 0.013$
    & $2.684 \pm 0.125$
    & $0.842 \pm 0.004$
    & $0.920 \pm 0.001$ \\
&
& OpDis-PS ({\bf ours})
    & ${\bf 0.321 \pm 0.001}^{\star}$
    & ${\bf 2.109 \pm 0.005}^{\star}$
    & ${\bf 2.470 \pm 0.090}^{\star}$
    & ${\bf 0.855 \pm 0.004}$
    & ${\bf 0.940 \pm 0.000}^{\star}$ \\
\bottomrule
\multicolumn{8}{l}{\footnotesize
RE = relative error; ATE = activation time error; LE = localization error; sCC = spatial corr. coefficient; tCC = temporal corr. coefficient}
\end{tabular}}
\end{table}

Tables~\ref{tab:phy_pat} and~\ref{tab:phy_healthy} report the pathological and physiological evaluations at 40 dB. OpDis-PS achieves the lowest RE and the highest sCC and tCC under both discrepancy sources on both datasets, with the lowest or near-lowest ATE and LE. Variational methods remain clearly less accurate than the leading data-driven methods. Amortized methods are competitive on physiological activation but, together with DPS, degrade most on pathological activation. Appendix Table~\ref{tab:nominal-physiological} provides the matched-operator physiological reference, Appendix~\ref{sec:additional-30db-results} reports the evaluations under stronger measurement noise, and Appendix~\ref{sec:ablation-covariance} ablates the discrepancy model.

{\bf Discussion.}
The results indicate that explicit operator information and a learned prior are complementary. Variational methods use the operator explicitly but rely on generic smoothness assumptions, which limits their accuracy. Amortized methods learn the observation process implicitly and perform well on activation patterns resembling the training data, but generalize less well to pathological activation. Posterior sampling unites both strengths, pairing an explicit operator with a learned prior, which makes the likelihood a direct lever for incorporating operator knowledge. Since OpDis-PS shares the prior, nominal operator, and sampler with DDRM, its consistently lower RE isolates the benefit of the discrepancy-aware likelihood.

\clearpage
\subsection{Qualitative insights}
\label{sec:qual_results}

\begin{figure}[t]
    \centering

    \begin{subfigure}[t]{0.25\textwidth}
        \centering
        \includegraphics[width=0.9\linewidth]{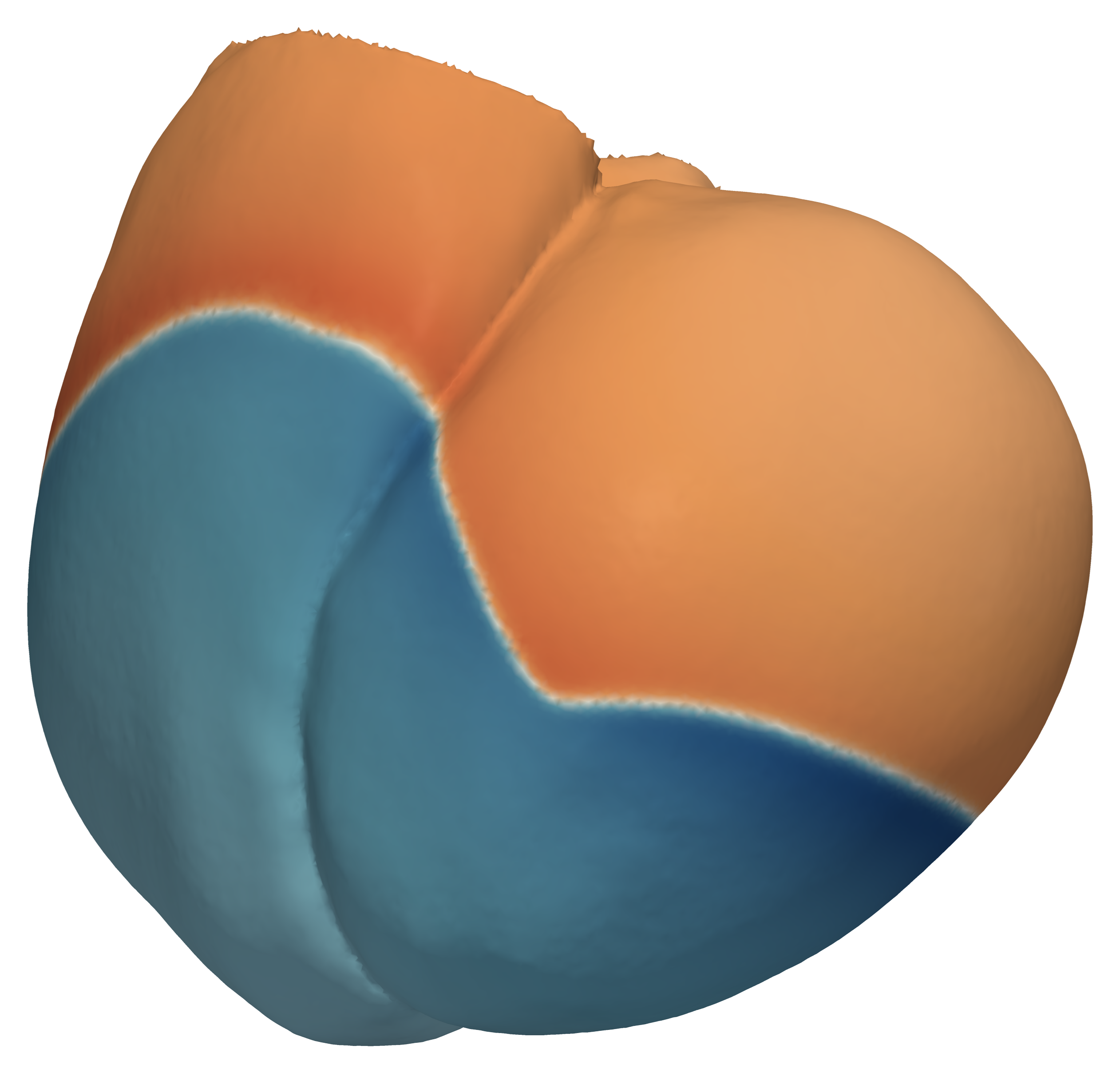}
        \caption{Ground truth.}
        \label{fig:sub-a}
    \end{subfigure}
    \qquad
    \begin{subfigure}[t]{0.25\textwidth}
        \centering
        \includegraphics[width=0.9\linewidth]{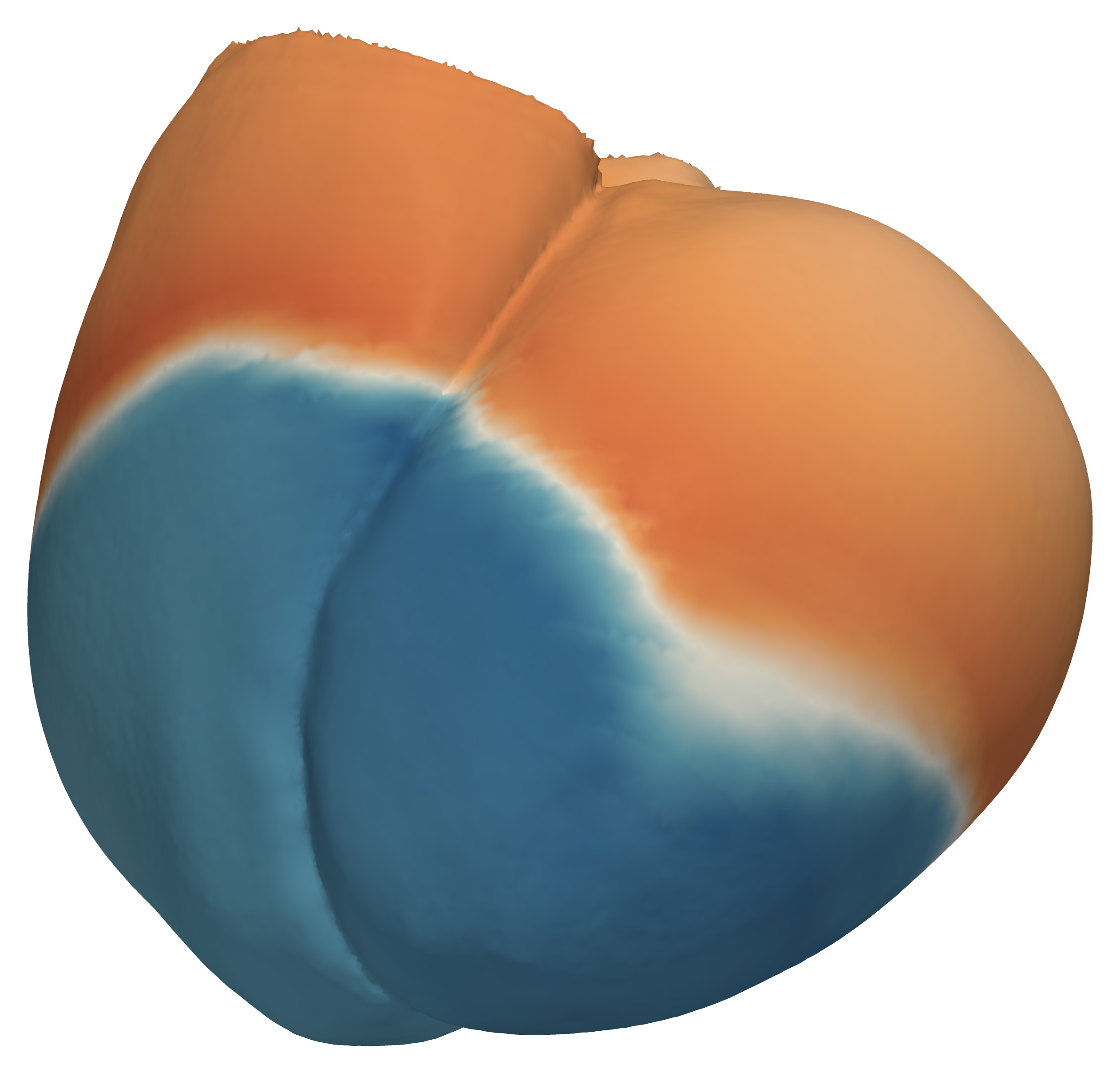}
        \caption{Palette.}
        \label{fig:sub-b}
    \end{subfigure}
    \qquad
    \begin{subfigure}[t]{0.25\textwidth}
        \centering
        \includegraphics[width=0.9\linewidth]{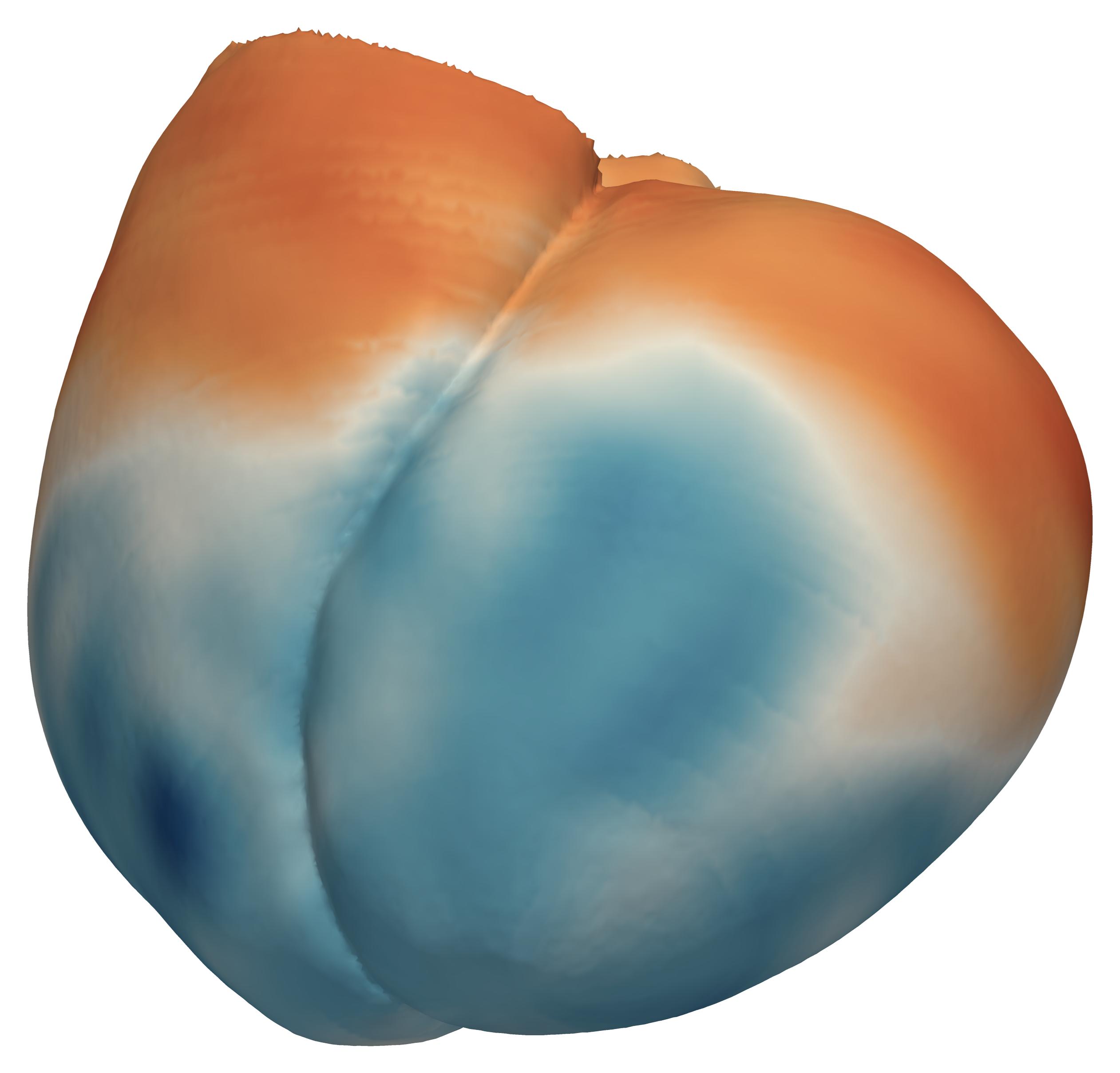}
        \caption{Transformer.}
        \label{fig:sub-c}
    \end{subfigure}

    \vspace{1.0em}

    \begin{subfigure}[t]{0.25\textwidth}
        \centering
        \includegraphics[width=0.9\linewidth]{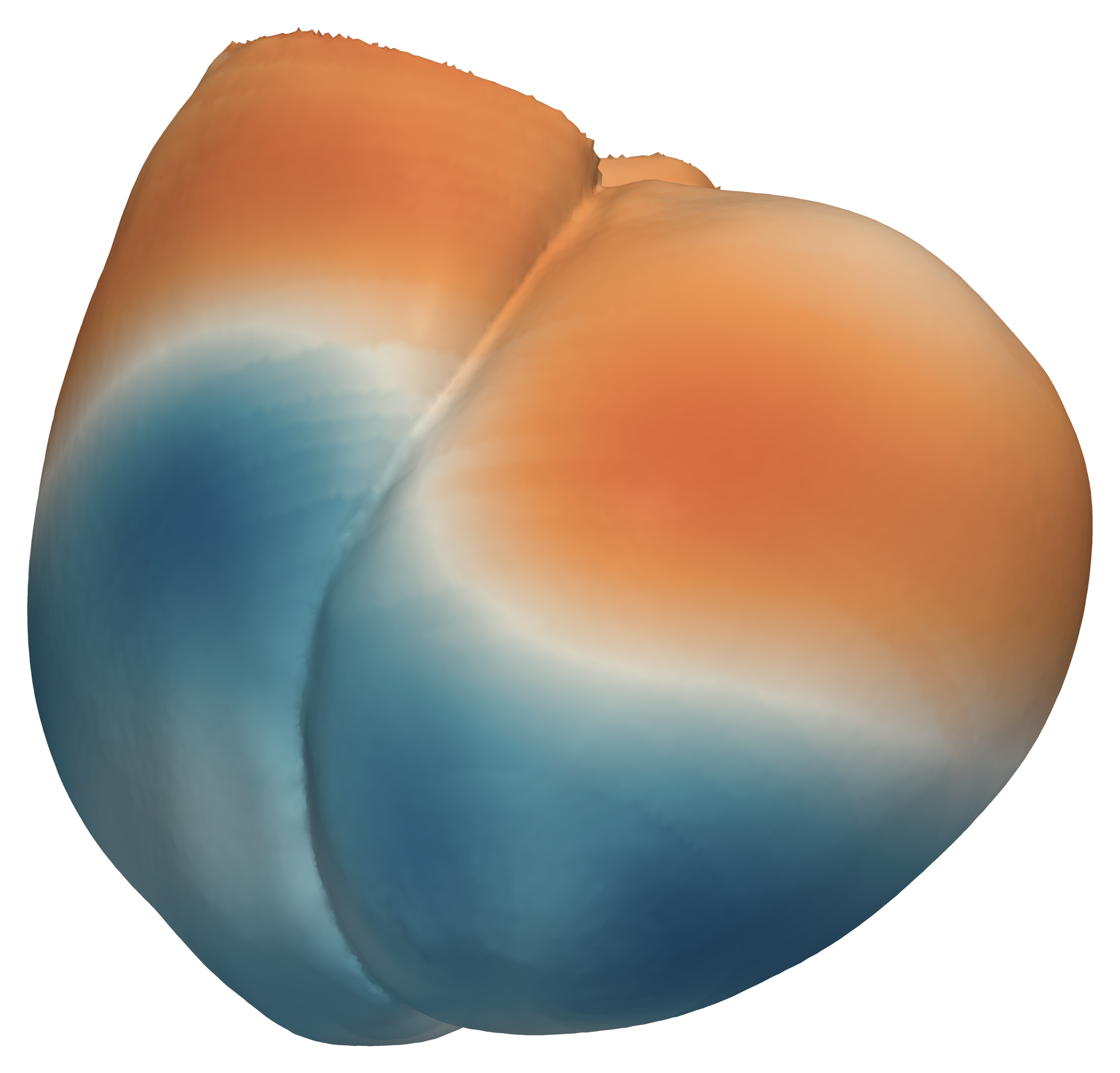}
        \caption{Tikhonov (spatial).}
        \label{fig:sub-d}
    \end{subfigure}
    \qquad
    \begin{subfigure}[t]{0.25\textwidth}
        \centering
        \includegraphics[width=0.9\linewidth]{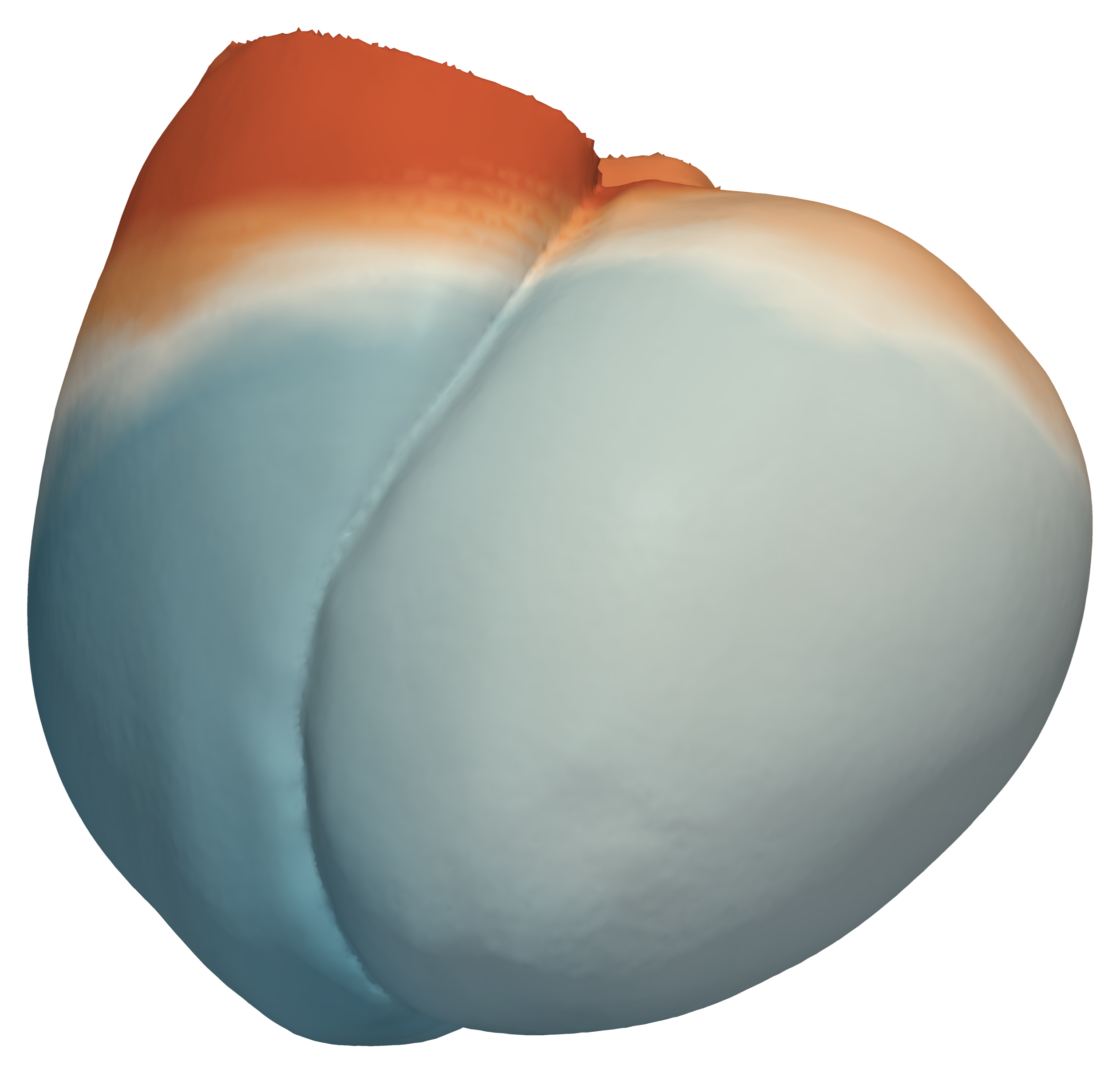}
        \caption{DPS.}
        \label{fig:sub-e}
    \end{subfigure}
    \qquad
    \begin{subfigure}[t]{0.25\textwidth}
        \centering
        \includegraphics[width=0.9\linewidth]{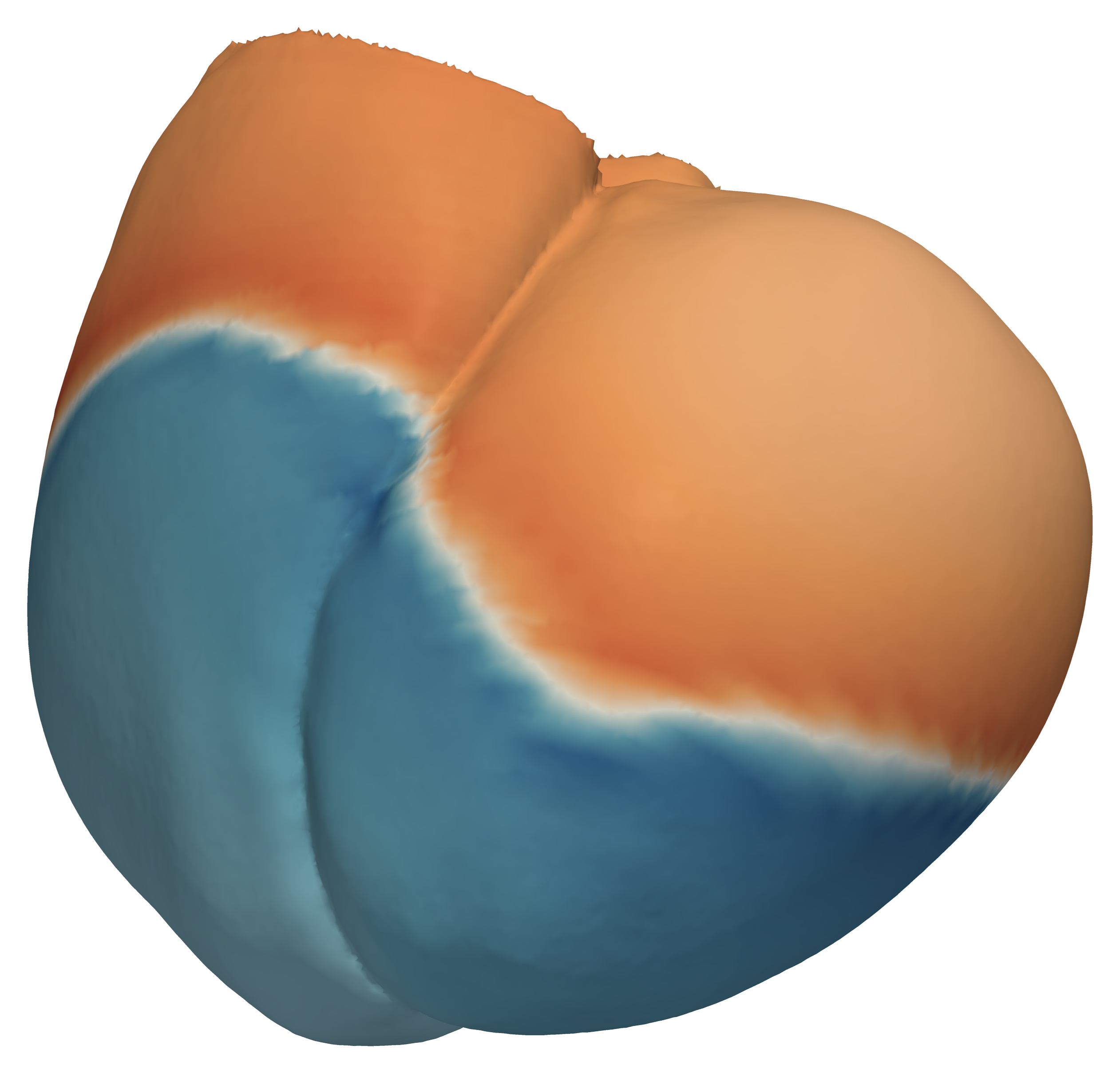}
        \caption{OpDis-PS ({\bf ours}).}
        \label{fig:sub-f}
    \end{subfigure}

    \caption{{\bf Qualitative insights.} Ground-truth and reconstructed heart-template potentials for an illustrative LBBB case under lung-conductivity discrepancy at \(t=110\,\mathrm{ms}\) and 40 dB. OpDis-PS and Palette reproduce the activation front, \mbox{which the Transformer blurs and DPS displaces.}}
    \label{fig:grid}
\end{figure}

Figure~\ref{fig:grid} compares heart-template potentials reconstructed for an illustrative LBBB case under lung-conductivity discrepancy at \(t=110\,\mathrm{ms}\) and 40 dB. Here, OpDis-PS and Palette recover the sharp activation front most clearly, with OpDis-PS showing the closest visual agreement with the ground truth. Spatial Tikhonov captures the characteristic LBBB-induced displacement of the front, but broadens it through spatial regularization. The Transformer yields a more diffuse front, whereas DPS exhibits a larger spatial displacement and attenuated contrast.

\section{Limitations}
\label{sec:limitations}

Our evaluation relies on simulated electrophysiology and two discrepancy sources. Real recordings may contain several sources of mismatch, including variations not represented in the operator distribution used to estimate the approximation-error model. In practice, this distribution would need to be constructed from plausible variation in anatomy, tissue conductivities, electrode placement, and acquisition conditions. In our experiments, it is narrower than the distribution generating the test operators (Section~\ref{sec:data}), but we have not systematically varied the degree of misspecification. The discrepancy model is linear-Gaussian and treats residuals as independent across time, although a recording shares one operator. Its linear predictor explains less of the lung-conductivity discrepancy than of the subcutaneous-fat discrepancy (held-out $R^2$ of $0.441$ versus $0.937$). Finally, the improvements are metric-dependent. Future work should assess their clinical relevance and uncertainty calibration on real recordings.
\section{Conclusion}
\label{sec:conclusion}

We introduced OpDis-PS, a diffusion posterior sampler for reconstruction under forward-operator discrepancy. A signal-conditional approximation-error model corrects the nominal operator and incorporates the remaining mismatch into a whitened likelihood, while leaving the unconditional diffusion prior fixed. Under both discrepancy sources at 40 dB, OpDis-PS achieves the lowest global reconstruction error and the highest spatial and temporal correlation, with its advantage over amortized baselines widening on pathological activation patterns excluded from training. These results show that adapting the likelihood to operator discrepancy can improve diffusion posterior sampling without retraining the generative prior.

\begin{ack}
The authors acknowledge the HPC resources provided by the Erlangen National High-Performance Computing Center (NHR@FAU) of the Friedrich-Alexander-Universität Erlangen-Nürnberg (FAU). We thank the developers of openCARP, the authors of the TotalSegmentator dataset, and the authors of the statistical shape model of \citet{Rodero2021} for making their resources publicly available.
\end{ack}
\bibliographystyle{plainnat}
\bibliography{references}

\appendix
\section{Mathematical details}
\label{sec:mathematical-details}

This appendix provides the mathematical details omitted from the main
text. We first derive the reduced average-referenced observation model
(Section~\ref{sec:coordinate-reduction}) and the signal-conditional
approximation-error likelihood
(Section~\ref{sec:signal-conditional-derivation}). We then describe the
heart-template construction (Section~\ref{sec:chart-construction}) and
provide precise definitions of the evaluation metrics
(Section~\ref{sec:metric-definitions}).

\subsection{Coordinate reduction}
\label{sec:coordinate-reduction}

The centering matrix $\mathbf C=\mathbf I-\frac{\mathbf 1\mathbf 1^{\mathsf T}}{n_{\mathrm B}}$ is symmetric and idempotent with rank $n_{\mathrm B}-1$, and $\mathbf C\mathbf 1=\mathbf 0$, so it removes any constant offset across the electrodes. The centered noise $\boldsymbol\varepsilon_t=\mathbf C\mathbf z_t$ then has the rank-deficient covariance $\sigma^2\mathbf C$. To recover an isotropic noise model, we choose $\mathbf R\in\mathbb R^{n_{\mathrm B}\times(n_{\mathrm B}-1)}$ whose columns form an orthonormal basis of the range of $\mathbf C$, so that
\begin{equation}
\mathbf R^{\mathsf T}\mathbf R=\mathbf I_{n_{\mathrm B}-1},
\qquad
\mathbf R\mathbf R^{\mathsf T}=\mathbf C.
\label{eq:R-properties}
\end{equation}
Left-multiplying the observation model $\mathbf b=\mathbf A\mathbf h+\boldsymbol\varepsilon$ by $\mathbf R^{\mathsf T}$ keeps the model linear,
\begin{equation}
\mathbf R^{\mathsf T}\mathbf b
=
\mathbf R^{\mathsf T}\mathbf A\,\mathbf h
+
\mathbf R^{\mathsf T}\boldsymbol\varepsilon,
\end{equation}
and by~\eqref{eq:R-properties} the noise covariance becomes isotropic,
\begin{equation}
\operatorname{Cov}(\mathbf R^{\mathsf T}\boldsymbol\varepsilon_t)
=
\sigma^2\mathbf R^{\mathsf T}\mathbf C\mathbf R
=
\sigma^2\mathbf R^{\mathsf T}(\mathbf R\mathbf R^{\mathsf T})\mathbf R
=
\sigma^2\mathbf I_{n_{\mathrm B}-1}.
\end{equation}
The reduced noise is therefore isotropic, as the spectral posterior sampler assumes. With slight abuse of notation, we reuse $\mathbf b$, $\mathbf A$, $\mathbf A_0$, and $\boldsymbol\varepsilon$ throughout the main text for the reduced quantities $\mathbf R^{\mathsf T}\mathbf b$, $\mathbf R^{\mathsf T}\mathbf A$, $\mathbf R^{\mathsf T}\mathbf A_0$, and $\mathbf R^{\mathsf T}\boldsymbol\varepsilon$.

\subsection{Signal-conditional approximation-error derivation}
\label{sec:signal-conditional-derivation}

We derive the correction of Section~\ref{sec:approximation-error-likelihood} for a single time point $t$. The standard conditional-Gaussian formula \citep{Bishop2006} applied to the joint model in Equation~\eqref{eq:joint-gaussian} gives
\begin{equation}
\mathbf d_t
=
\boldsymbol\mu_{\mathrm d}
+
\mathbf L(\mathbf s_t-\boldsymbol\mu_{\mathrm s})
+
\mathbf r_t,
\qquad
\mathbf r_t
\sim
\mathcal N(\mathbf 0,\boldsymbol\Sigma_{\mathrm r}),
\end{equation}
where
\begin{equation}
\mathbf L
=
\mathbf C_{\mathrm{ds}}\mathbf C_{\mathrm{ss}}^{-1},
\qquad
\boldsymbol\Sigma_{\mathrm r}
=
\mathbf C_{\mathrm{dd}}-\mathbf L\mathbf C_{\mathrm{sd}}.
\end{equation}
Under the joint-Gaussian model, the residual distribution does not depend on the nominal signal $\mathbf s_t$.

Substituting $\mathbf s_t=\mathbf A\mathbf h_t$ into the reduced observation model $\mathbf b_t=\mathbf A\mathbf h_t+\mathbf d_t+\boldsymbol\varepsilon_t$ gives
\begin{align}
\mathbf b_t
&=
\mathbf A\mathbf h_t
+
\boldsymbol\mu_{\mathrm d}
+
\mathbf L(\mathbf A\mathbf h_t-\boldsymbol\mu_{\mathrm s})
+
\mathbf r_t
+
\boldsymbol\varepsilon_t
\\
&=
(\mathbf I+\mathbf L)\mathbf A\mathbf h_t
+
(\boldsymbol\mu_{\mathrm d}-\mathbf L\boldsymbol\mu_{\mathrm s})
+
\mathbf r_t
+
\boldsymbol\varepsilon_t.
\end{align}
Collecting the signal-dependent part into a corrected operator and the constant part into a corrected observation,
\begin{equation}
\tilde{\mathbf A}
=
(\mathbf I+\mathbf L)\mathbf A,
\qquad
\tilde{\mathbf b}_t
=
\mathbf b_t
-
(\boldsymbol\mu_{\mathrm d}-\mathbf L\boldsymbol\mu_{\mathrm s}),
\end{equation}
the corrected model reads
\begin{equation}
\tilde{\mathbf b}_t
=
\tilde{\mathbf A}\mathbf h_t
+
\tilde{\boldsymbol\varepsilon}_t,
\qquad
\tilde{\boldsymbol\varepsilon}_t
=
\mathbf r_t+\boldsymbol\varepsilon_t.
\end{equation}
Since the measurement noise $\boldsymbol\varepsilon_t\sim \mathcal N(\mathbf 0,\sigma^2\mathbf I_{n_{\mathrm B}-1})$ is independent of the residual discrepancy $\mathbf r_t$, their covariances add, yielding
\begin{equation}
\tilde{\boldsymbol\varepsilon}_t
\sim
\mathcal N\!\left(
\mathbf 0,
\boldsymbol\Sigma_{\mathrm r}
+
\sigma^2\mathbf I_{n_{\mathrm B}-1}
\right).
\end{equation}
We therefore define the whitening matrix
\begin{equation}
\mathbf W
=
\left(
\boldsymbol\Sigma_{\mathrm r}
+
\sigma^2\mathbf I_{n_{\mathrm B}-1}
\right)^{-1/2},
\end{equation}
which maps the combined noise covariance to the identity. We estimate $\mathbf L$ and $\boldsymbol\Sigma_{\mathrm r}$ from finite
training data as described in
Appendix~\ref{sec:opdis-discrepancy-estimation}. Applying the correction
and whitening column-wise yields the sequence-level model in
Equation~\eqref{eq:conditional-whitening}.

\subsection{Template-coordinate construction and surface mappings}
\label{sec:chart-construction}

We construct the template coordinates on the heart surface $\Gamma_{\mathrm H}$ of one reference heart. To localize the apex, we solve the surface Poisson problem
\begin{equation}
-\Delta_{\Gamma_{\mathrm H}}w=1,
\qquad
w=0
\quad\text{on the valve rims},
\label{eq:apex-localization}
\end{equation}
whose solution $w$ is a smooth intrinsic apex indicator. Its maximum defines the apex location, and the $3\%$ of surface area with the largest values of $w$ forms a finite apical cap.

Between this cap and the valve rims, we then solve the Laplace--Beltrami problem
\begin{equation}
-\Delta_{\Gamma_{\mathrm H}}u=0,
\qquad
u=0
\quad\text{on the apical cap},
\qquad
u=1
\quad\text{on the valve rims},
\label{eq:apicobasal-harmonic-function}
\end{equation}
whose harmonic solution $u$ varies smoothly from apex to base. We discard the apical cap and the basal collar with $u>u_{\mathrm{cut}}$, where $u_{\mathrm{cut}}=0.99$, and denote the remaining surface by $\Gamma_{\mathrm{rep}}$.

Cumulative surface area across the level sets of $u$ defines the apico-basal coordinate $z_h$ and divides $\Gamma_{\mathrm{rep}}$ into $30$ approximately equal-area bands. Within each band, cumulative surface area defines the periodic circumferential coordinate $\varphi_h$ with $60$ positions, and we align the circumferential origins of consecutive bands through their nearest surface points. This produces the regular $30\times60$ grid of heart-template locations, which we transfer to every other heart through the known node correspondence.

Let $\mathbf M_{\mathrm H}$ be the lumped heart-mesh mass matrix, with weights outside $\Gamma_{\mathrm{rep}}$ set to zero. The mass-weighted $L^2$ fit of the heart-mesh potentials $\Phi_{\mathrm H}$ is
\begin{equation}
\mathbf h
=
\left(
\mathbf T_{\mathrm H}^{\mathsf T}
\mathbf M_{\mathrm H}
\mathbf T_{\mathrm H}
\right)^{-1}
\mathbf T_{\mathrm H}^{\mathsf T}
\mathbf M_{\mathrm H}
\Phi_{\mathrm H}.
\label{eq:template-fit}
\end{equation}
The same map $\mathbf T_{\mathrm H}$ interpolates heart-template potentials back to the heart-mesh nodes.

\subsection{Metrics}
\label{sec:metric-definitions}

Let $\widehat{\mathbf h},\mathbf h\in\mathbb R^{n_{\mathrm H}\times T}$ denote the reconstructed and ground-truth heart-template potentials, with $n_{\mathrm H}=1800$. The entry $h_{i,t}$ is the potential at heart-template location $i$ and time point $t$, and consecutive time points are separated by $5$~ms.

{\bf Relative error (RE).}
The global Frobenius relative error is given by
\begin{equation}
\operatorname{RE}
=
\frac{\lVert \widehat{\mathbf h}-\mathbf h\rVert_F}{\lVert\mathbf h\rVert_F}.
\label{eq:re-definition}
\end{equation}

{\bf Spatial and temporal correlation coefficients (sCC, tCC).}
We compute sCC as the mean spatial Pearson correlation over time,
\begin{equation}
\operatorname{sCC}
=
\frac{1}{|\mathcal T|}
\sum_{t\in\mathcal T}
\frac{
\sum_{i=1}^{n_{\mathrm H}}
(\widehat h_{i,t}-\widehat\mu_t)
(h_{i,t}-\mu_t)
}{
\sqrt{
\sum_{i=1}^{n_{\mathrm H}}(\widehat h_{i,t}-\widehat\mu_t)^2
\sum_{i=1}^{n_{\mathrm H}}(h_{i,t}-\mu_t)^2
}
}.
\end{equation}
Here $\mu_t=n_{\mathrm H}^{-1}\sum_{i=1}^{n_{\mathrm H}}h_{i,t}$ is the spatial mean of the ground truth and $\widehat\mu_t$ the corresponding reconstruction mean. The set $\mathcal T$ collects the time points with nondegenerate spatial variance in both the reconstruction and the ground truth. For tCC, we instead compute the Pearson correlation between reconstructed and true waveforms at each heart-template location and average over locations with nondegenerate temporal variance.

{\bf Activation time error (ATE).}
We estimate the temporal derivative $\dot h_{i,t}$ by finite differences and take the local activation time as the time of the steepest downstroke,
\begin{equation}
\mathrm{LAT}_i(\mathbf h)
=
(5\,\mathrm{ms})
\left(
\operatorname*{arg\,max}_{t\in\{1,\ldots,T\}}
(-\dot h_{i,t})-1
\right),
\end{equation}
where the trailing $-1$ places the first time point at $0$~ms. ATE is the mean absolute activation-time error in milliseconds,
\begin{equation}
\operatorname{ATE}
=
\frac{1}{n_{\mathrm H}}\sum_{i=1}^{n_{\mathrm H}}
\bigl|\mathrm{LAT}_i(\widehat{\mathbf h})-\mathrm{LAT}_i(\mathbf h)\bigr|.
\end{equation}

{\bf Localization error (LE).}
We call a heart-template location amplitude-gated if its temporal peak-to-peak amplitude reaches $30\%$ of the maximum across locations. An early-breakthrough region is a connected cluster (link radius $12$~mm) of gated locations whose LAT lies within $5$~ms of the earliest gated LAT, and each region contributes its centroid in millimeters. Because LATs are resolved on the $5$~ms sampling grid, several reconstructed regions can share the earliest gated LAT. Let $\{\mathbf c_{\widehat{\mathbf h}}^{(j)}\}_{j=1}^{J}$ denote the centroids of the reconstructed regions that contain a location at the earliest gated LAT, and $\{\mathbf c_{\mathbf h}^{(k)}\}$ the set of ground-truth centroids. The localization error is
\begin{equation}
\operatorname{LE}
=
\frac{1}{J}\sum_{j=1}^{J}\min_k\bigl\lVert
\mathbf c_{\widehat{\mathbf h}}^{(j)}-\mathbf c_{\mathbf h}^{(k)}
\bigr\rVert_2.
\end{equation}
The one-to-nearest convention accounts for near-simultaneous fronts on the true heart surface, and averaging over the $J$ earliest reconstructed regions makes LE independent of the order of tied regions.

{\bf Statistical testing.}
Data-driven methods are trained five times. DPS, DDRM, and OpDis-PS use the same five trained diffusion models, so their results pair run by run. For each metric and condition, we compare the best and second-best data-driven methods with a two-sided paired $t$-test when both use the same trained models and with a two-sided Welch $t$-test otherwise.
\section{Architectural details}
\label{sec:architectural-details}

We describe the Transformer and Palette architectures used for amortized inference (Sections~\ref{sec:transformer-architecture} and~\ref{sec:conditional-diffusion-architecture}), followed by the unconditional diffusion prior shared by all posterior-sampling methods (Section~\ref{sec:unconditional-prior-architecture}).
All data-driven architectures are implemented in PyTorch.

\subsection{Amortized inference}

\subsubsection{Transformer}
\label{sec:transformer-architecture}

The Transformer baseline is an encoder-only self-attention model~\citep{Vaswani2017} applied along the time axis. Each of the 91 time points forms a token containing the 176 centered body-template potentials $\mathbf b$ (without the coordinate reduction of Appendix~\ref{sec:coordinate-reduction}). A linear layer projects each token to a hidden dimension of 384, and fixed sinusoidal positional encodings inject its temporal position.

The encoder uses PyTorch's \texttt{TransformerEncoder} with six pre-LayerNorm blocks (\texttt{norm\_first=True}), each with eight attention heads, a feed-forward width of 1536, GELU activations, and a dropout probability of $0.1$. The encoder processes the full sequence without a causal mask. A shared linear head maps each output token to the 1800 heart-template potentials $\mathbf h$ at the corresponding time point. The network has $11{,}408{,}520$ trainable parameters.

\subsubsection{Palette}
\label{sec:conditional-diffusion-architecture}

We employ a Palette-style conditional VP-DDPM~\citep{Saharia2022Palette} that maps body-template potentials to heart-template potentials, in line with the geometry-free conditional-diffusion approach to ECGI of~\citet{JaraMeyers2026}. It supplies a fixed condition alongside the noisy target at every denoising step, with a 3D U-Net backbone and physical boundary conditions.

At each time point, a shared linear layer maps the 176 centered body-template potentials to 1800 conditioning values, reshaped into a $30\times60$ grid aligned with the heart-template locations. This places the observation in the reconstruction's own coordinate system, and we train the projection jointly with the diffusion model.

The conditional denoiser shares the 3D U-Net architecture of the unconditional prior (Section~\ref{sec:unconditional-prior-architecture}) with separately trained weights. Its six input channels stack the diffusion-corrupted heart-template potentials, the conditioning grid, and the four fixed coordinate channels. The complete model has $15{,}205{,}705$ trainable parameters, including $318{,}600$ in the conditioning projection.

\subsection{Posterior sampling}

\subsubsection{Unconditional diffusion prior}
\label{sec:unconditional-prior-architecture}

The three posterior samplers, OpDis-PS, DDRM, and DPS, reconstruct with the same frozen unconditional prior and differ only in the likelihood. The prior operates on complete heart-template potentials on the fixed $91\times30\times60$ grid over time and heart-template locations. Its denoiser has five input channels containing the diffusion-corrupted heart-template potentials and the fixed coordinates $z_{h}$, $\sin(\varphi_{h})$, $\cos(\varphi_{h})$, and normalized time. Its single output channel predicts the diffusion velocity $\mathbf v_\theta$~\citep{SalimansHo2022}.

The denoiser is a 3D U-Net that jointly processes time and both template axes. The input tensor is padded to $96\times32\times60$ before the U-Net and cropped back afterward. The $3\times3\times3$ convolutions use replicated boundaries in time and along $z_{h}$ and a circular boundary along $\varphi_{h}$.

The encoder levels and bottleneck have widths 32, 64, 128, and 256, with two residual blocks at each level. The three downsampling strides are $(2,2,2)$, $(2,2,2)$, and $(2,2,1)$. The decoder uses corresponding nearest-neighbor upsampling followed by convolutions and concatenates encoder features through skip connections. Each decoder level contains two residual blocks.

The residual blocks use GroupNorm with eight groups, SiLU activations, and a dropout probability of $0.1$. A sinusoidal diffusion-step embedding followed by a two-layer MLP enters every residual block. The network contains no attention layers and has $14{,}886{,}241$ trainable parameters.

\section{Hyperparameters}
\label{sec:hyperparameters}

We summarize the training and inference settings of all methods and describe the validation procedures used for hyperparameter selection. We then detail the estimation of the OpDis-PS discrepancy model (Section~\ref{sec:opdis-discrepancy-estimation}).

\subsection{Variational regularization}

{\bf First-order Tikhonov.}
We compare spatial regularization orders zero, one, and two on nominal validation and first-order regularization achieves the lowest mean global relative error. We use $\mathbf M_{\mathrm G}= \mathbf L_{\mathrm G}+\gamma\mathbf I$, where $\mathbf L_{\mathrm G}$ is the unweighted four-neighbor graph Laplacian of the heart-template locations. We set $\gamma=10^{-3}\lambda_{\min}^{+}(\mathbf L_{\mathrm G})$, with $\lambda_{\min}^{+}$ denoting the smallest positive eigenvalue. GCV selects one $\lambda$ per case from 1000 logarithmically spaced values between $10^{-8}$ and $10^{2}$ using all time points. We retain eigenmodes of $\mathbf A\mathbf M_{\mathrm G}^{-1}\mathbf A^{\mathsf T}$ with $w_i>10^{-6}w_{\max}$.

{\bf Spatio-temporal Tikhonov.}
Spatio-temporal Tikhonov uses the same spatial penalty and effective-rank cutoff. It adds a second-order temporal-difference penalty and fixes $\rho=\lambda_{\mathrm t}^2/\lambda_{\mathrm s}^2=1$. Joint GCV selects the common scale from the same grid for each reconstruction.

{\bf Total Variation.}
We select one SNR-independent strength from $\{0.001,0.003,0.01,0.03,0.1,0.3\}$ by mean global relative error across the nominal validation conditions. This yields $\alpha=0.01$. The solver uses a quadratic gauge-fixing term with $\epsilon=10^{-6}$, a relative-update tolerance of $10^{-5}$, and at most $100{,}000$ iterations.

\subsection{Amortized inference}

{\bf Transformer.}
We train five independently initialized models on the 5100 nominal training pairs. Body-template potentials and heart-template potentials use separate scalar standard deviations estimated from the training set. During training, we augment each body-template input with probability $0.8$ using centered Gaussian noise whose target SNR is drawn uniformly in dB from $[10,40]$. We use AdamW with batch size 8, learning rate $10^{-4}$, weight decay $10^{-4}$, and gradient-norm clipping at 1. Training runs for at most 100 epochs and stops after 15 consecutive epochs without lower validation MSE. We select the checkpoint with the lowest MSE on the 100 nominal validation pairs. We fix the architecture of Appendix~\ref{sec:transformer-architecture} without a validation sweep.

{\bf Palette.}
We train five independently initialized models on the same training pairs and use the same normalization and augmentation protocol. Each model uses a 1000-step cosine variance-preserving diffusion process with the $\mathbf v$-parameterization \citep{NicholDhariwal2021,SalimansHo2022}. We train for 150 epochs with AdamW using batch size 4, learning rate $2\times10^{-4}$, weight decay $10^{-4}$, gradient-norm clipping at 1, and EMA decay $0.9999$. We select one checkpoint per training by fixed-seed $\mathbf v$-MSE on the same 100 nominal validation pairs and use its EMA weights at inference. Sampling uses 50 DDIM steps with $\eta=0$, temperature 1, and base inference seed 0. We report the mean of 16 samples as the point estimate.

\subsection{Posterior sampling}

{\bf Unconditional diffusion prior.}
We train five independently initialized priors only on the 5100 physiological heart-template potentials, normalized by their global training-set standard deviation. Each prior uses a 1000-step cosine variance-preserving diffusion process with the $\mathbf v$-parameterization \citep{NicholDhariwal2021,SalimansHo2022}. We train for 54 epochs with AdamW using batch size 4, learning rate $2\times10^{-4}$, weight decay $10^{-4}$, and EMA decay $0.9999$. Unconditional prior evaluation on validation supports the use of $\mathbf v$-parameterization and EMA weights. Since the late EMA checkpoints perform similarly, all posterior-sampling methods use the checkpoint after the final training epoch.

\begin{table}[t]
\centering
\caption{{\bf DDRM sampler selection.}
Mean global relative error on 100 nominal validation cases at 40 dB using one trained diffusion model. Bold marks the best parameter combination.}
\label{tab:ddrm-sampler-selection}
\begin{tabular}{cccc}
\toprule
\multirow{2}{*}{$\eta$} & \multicolumn{3}{c}{Mean RE\,$\downarrow$} \\
\cmidrule(lr){2-4}
& $\eta_b=0.5$ & $\eta_b=0.75$ & $\eta_b=1$ \\
\midrule
$0.85$ & $0.3719$ & $0.3076$ & $0.2688$ \\
$1.0$  & $0.3903$ & $0.3088$ & $\mathbf{0.2609}$ \\
\bottomrule
\end{tabular}
\end{table}

{\bf DDRM and OpDis-PS.}
DDRM has two stochasticity hyperparameters: $\eta_b$ controls how strongly the update moves towards the observation in singular directions where the iterate is noisier than the observation, and $\eta$ controls the weak measurement guidance in the remaining directions. With $\eta_b=1$, noiseless observations are enforced exactly in every observed coordinate.
DDRM and OpDis-PS use the same sampler settings. On nominal validation at 40 dB with 100 cases and one of the five independently trained priors, we jointly compare $\eta\in\{0.85,1\}$ and $\eta_b\in\{0.5,0.75,1\}$. Table~\ref{tab:ddrm-sampler-selection} reports the mean global relative errors. The combination $\eta=\eta_b=1$ achieves the lowest error among these six configurations. We use this setting across the 30- and 40-dB evaluation conditions and all operator conditions. Both methods use 20 reverse steps and report the mean of 16 samples as the point estimate.

{\bf DPS.}
We select the guidance strength $\zeta$ on nominal validation at 20 dB using 100 cases and one of the five independently trained priors. We compare $\zeta\in\{0,10,40,60,100,150\}$ while fixing 50 DDIM steps, $\eta=0$, temperature 1, and the point estimate to the mean of 16 samples. The corresponding mean global relative errors are $0.6925$, $0.5248$, $0.4820$, $0.4921$, $0.5169$, and $0.5417$. We therefore select $\zeta=40$.

\subsection{OpDis-PS discrepancy estimation}
\label{sec:opdis-discrepancy-estimation}

We fit a separate discrepancy model for each discrepancy source on 64 training heart--torso pairs, each combined with six recording operators $\mathbf A_0$ drawn from the corresponding perturbation distribution (Section~\ref{sec:data}), so that a single set of estimates $(\widehat{\boldsymbol\mu}_{\mathrm s},\widehat{\boldsymbol\mu}_{\mathrm d},\widehat{\mathbf L},\widehat{\boldsymbol\Sigma}_{\mathrm r})$ per source is shared across all test geometries.

{\bf Signal-conditional discrepancy estimator.}
 For each training time point in the $p=n_{\mathrm B}-1=175$-dimensional average-referenced measurement space, we form $\mathbf s_t=\mathbf A\mathbf h_t$ and $\mathbf d_t=(\mathbf A_0-\mathbf A)\mathbf h_t$, yielding 34{,}944 pairs per source. We estimate the conditional-mean predictor as
\begin{equation}
\widehat{\mathbf L}
=
\widehat{\mathbf C}_{\mathrm{ds}}
\left(\widehat{\mathbf C}_{\mathrm{ss}}+\lambda\mathbf I\right)^{-1},
\qquad
\lambda
=
\kappa\frac{\operatorname{tr}(\widehat{\mathbf C}_{\mathrm{ss}})}{p},
\end{equation}
where $\widehat{\mathbf C}_{\mathrm{ss}}$ and $\widehat{\mathbf C}_{\mathrm{ds}}$ are the sample covariance blocks of $(\mathbf s_t,\mathbf d_t)$. Geometry-level 8-fold cross-validation over $\kappa\in\{0,10^{-4},10^{-3},10^{-2},10^{-1},1,10\}$ selects $\kappa=10^{-3}$ for both sources by held-out coefficient of determination. The selected predictor achieves held-out $R^2$ values of $0.441$ for lung conductivity and $0.937$ for subcutaneous fat. We keep complete geometries within a single fold and never split individual time points across training and validation.

We recompute the residuals $\mathbf r_t=\mathbf d_t-\widehat{\boldsymbol\mu}_{\mathrm d}-\widehat{\mathbf L}(\mathbf s_t-\widehat{\boldsymbol\mu}_{\mathrm s})$ and estimate their covariance using Ledoit--Wolf shrinkage \citep{LedoitWolf2004},
\begin{equation}
\widehat{\boldsymbol\Sigma}_{\mathrm r}
=
(1-\rho) \widehat{\mathbf C}_{\mathrm r \mathrm r}
+
\rho\frac{\operatorname{tr}(\widehat{\mathbf C}_{\mathrm r \mathrm r})}{p}\mathbf I.
\end{equation}
The analytic Ledoit--Wolf rule determines $\rho$ from the training residuals without an additional validation hyperparameter, yielding $\rho\approx7.2\times10^{-4}$ for lung conductivity and $3.8\times10^{-4}$ for subcutaneous fat.

{\bf Marginal discrepancy estimator.}
For the marginal ablations in Appendix~\ref{sec:ablation-covariance}, we estimate the discrepancy mean $\widehat{\boldsymbol\mu}_{\mathrm d}$ and covariance $\widehat{\mathbf C}_{\mathrm{dd}}$ directly from the discrepancy samples, applying the same automatic Ledoit--Wolf shrinkage to the sample covariance.

\section{Additional quantitative results}
\label{sec:additional-quantitative-results}

We first compare the magnitudes of measurement noise and operator discrepancy (Section~\ref{sec:discrepancy-magnitude}). We then report a matched-operator reference (Section~\ref{sec:nominal-reference}) and additional 30-dB results (Section~\ref{sec:additional-30db-results}), followed by ablations (Section~\ref{sec:ablation-covariance}) and an analysis of inference cost (Section~\ref{sec:inference-cost}).

\subsection{Noise and operator-discrepancy magnitude}
\label{sec:discrepancy-magnitude}

To compare the magnitude of operator discrepancy with measurement noise, we compute
\begin{equation}
r_{\mathrm{op}}
=
\frac{\lVert(\mathbf A_0-\mathbf A)\mathbf h\rVert_F}
     {\lVert\mathbf A\mathbf h\rVert_F}
\end{equation}
before adding measurement noise. Across the 250 physiological and 720 pathological evaluation cases, the median relative discrepancy is $5.4\%$ (interquartile range $2.4$--$8.1\%$) for lung conductivity and $15.2\%$ ($9.1$--$21.8\%$) for subcutaneous fat. Under the same norm convention, these values correspond to effective levels of $25.3$ and $16.3$ dB, respectively. For comparison, target SNRs of 40 and 30 dB correspond to relative noise norms of $1.0\%$ and $3.2\%$. Thus, the 40-dB condition emphasizes operator discrepancy without it being masked by measurement noise.

\subsection{Nominal reference}
\label{sec:nominal-reference}

Table~\ref{tab:nominal-physiological} reports the physiological results at 40 dB when the
recording and reconstruction operators coincide, $\mathbf A_0=\mathbf A$. We omit OpDis-PS because
no operator discrepancy needs to be modeled. Comparing these results with
Table~\ref{tab:phy_healthy} shows the degradation caused by operator discrepancy.

\begin{table}[t]
\centering
\caption{{\bf Nominal quantitative results (physiological data; 40 dB).} We evaluate seven baseline methods on the same 250 physiological test cases. For data-driven methods, values show the mean and standard deviation over five training runs.}
\label{tab:nominal-physiological}
\resizebox{\textwidth}{!}{%
\begin{tabular}{clccccc}
\toprule
& \multirow{2}{*}{Method} & \multicolumn{5}{c}{Metric} \\
\cmidrule(lr){3-7}
& & RE\,$\downarrow$ & ATE [ms]\,$\downarrow$ & LE [mm]\,$\downarrow$ & sCC\,$\uparrow$ & tCC\,$\uparrow$ \\
\midrule
\multirow{3}{*}{R}
& Tikhonov (spatial)
    & $0.518$
    & $5.507$
    & $15.574$
    & $0.661$
    & $0.827$ \\
& Tikhonov (spatio-temporal)
    & $0.506$
    & $4.715$
    & $8.434$
    & $0.687$
    & $0.835$ \\
& Total Variation
    & $0.573$
    & $6.058$
    & $9.869$
    & $0.743$
    & $0.805$ \\
\cmidrule(lr){1-7}
\multirow{2}{*}{A}
& Transformer
    & $0.392 \pm 0.002$
    & $2.796 \pm 0.048$
    & $4.561 \pm 0.294$
    & $0.814 \pm 0.009$
    & $0.906 \pm 0.001$ \\
& Palette
    & $0.377 \pm 0.002$
    & $2.709 \pm 0.021$
    & $5.035 \pm 0.194$
    & $0.852 \pm 0.003$
    & $0.912 \pm 0.001$ \\
\cmidrule(lr){1-7}
\multirow{2}{*}{P}
& DPS
    & $0.472 \pm 0.004$
    & $4.366 \pm 0.091$
    & $8.163 \pm 0.369$
    & $0.801 \pm 0.006$
    & $0.858 \pm 0.003$ \\
& DDRM
    & ${\bf 0.264 \pm 0.001}$
    & ${\bf 1.687 \pm 0.007}$
    & ${\bf 1.939 \pm 0.065}$
    & ${\bf 0.868 \pm 0.002}$
    & ${\bf 0.959 \pm 0.000}$ \\
\bottomrule
\multicolumn{7}{l}{\footnotesize
RE = relative error; ATE = activation time error; LE = localization error; sCC = spatial corr. coefficient; tCC = temporal corr. coefficient}
\end{tabular}}
\end{table}

\subsection{Additional results at 30 dB}
\label{sec:additional-30db-results}

Tables~\ref{tab:physiological-30db} and~\ref{tab:pathological-30db} report the corresponding evaluations at 30 dB, where measurement noise is more pronounced relative to the operator discrepancy than in the 40-dB condition.

\begin{table}[t]
\centering
\caption{{\bf Quantitative results (physiological data; 30 dB).}
We evaluate all methods on 250 physiological test cases under lung-conductivity and subcutaneous-fat discrepancy. Data-driven results are mean $\pm$ standard deviation over five training runs; $\star$ denotes $p<0.05$ for the best versus second-best data-driven method (Appendix~\ref{sec:metric-definitions}).}
\label{tab:physiological-30db}
\resizebox{\textwidth}{!}{%
\begin{tabular}{cclccccc}
\toprule
\multirow{2}{*}{Discrepancy}
& & \multirow{2}{*}{Method}
& \multicolumn{5}{c}{Metric} \\
\cmidrule(lr){4-8}
& & & RE\,$\downarrow$
& ATE [ms]\,$\downarrow$
& LE [mm]\,$\downarrow$
& sCC\,$\uparrow$
& tCC\,$\uparrow$ \\
\midrule

\multirow{8}{*}{\rotatebox[origin=c]{90}{\shortstack{lung\\conductivity}}}
& \multirow{3}{*}{R}
& Tikhonov (spatial)
& $0.701$
& $15.968$
& $39.565$
& $0.569$
& $0.752$ \\
&
& Tikhonov (spatio-temporal)
& $0.687$
& $6.896$
& $10.128$
& $0.608$
& $0.769$ \\
&
& Total Variation
& $0.584$
& $6.315$
& $10.373$
& $0.675$
& $0.795$ \\
\cmidrule(lr){2-8}
&
\multirow{2}{*}{A}
& Transformer
& $0.395 \pm 0.002$
& $2.822 \pm 0.046$
& $4.362 \pm 0.192$
& $0.814 \pm 0.009$
& $0.905 \pm 0.001$ \\
&
& Palette
& $0.380 \pm 0.003$
& $2.743 \pm 0.028$
& $5.360 \pm 0.268$
& ${\bf 0.851 \pm 0.002}$
& $0.910 \pm 0.002$ \\
\cmidrule(lr){2-8}
&
\multirow{3}{*}{P}
& DPS
& $0.483 \pm 0.004$
& $4.482 \pm 0.088$
& $8.695 \pm 0.343$
& $0.795 \pm 0.005$
& $0.852 \pm 0.003$ \\
&
& DDRM
& $0.363 \pm 0.001$
& $2.486 \pm 0.006$
& ${\bf 2.940 \pm 0.118}$
& $0.841 \pm 0.002$
& $0.923 \pm 0.000$ \\
&
& OpDis-PS ({\bf ours})
& ${\bf 0.347 \pm 0.001}^{\star}$
& ${\bf 2.426 \pm 0.010}^{\star}$
& $2.957 \pm 0.170$
& $0.849 \pm 0.003$
& ${\bf 0.927 \pm 0.000}^{\star}$ \\

\midrule

\multirow{8}{*}{\rotatebox[origin=c]{90}{\shortstack{subcutaneous\\fat}}}
& \multirow{3}{*}{R}
& Tikhonov (spatial)
& $0.697$
& $14.876$
& $40.930$
& $0.568$
& $0.756$ \\
&
& Tikhonov (spatio-temporal)
& $0.679$
& $6.417$
& $12.460$
& $0.608$
& $0.771$ \\
&
& Total Variation
& $0.602$
& $6.328$
& $9.460$
& $0.668$
& $0.794$ \\
\cmidrule(lr){2-8}
&
\multirow{2}{*}{A}
& Transformer
& $0.399 \pm 0.004$
& $2.882 \pm 0.056$
& $4.564 \pm 0.257$
& $0.810 \pm 0.008$
& $0.903 \pm 0.002$ \\
&
& Palette
& $0.376 \pm 0.003$
& $2.701 \pm 0.025$
& $4.984 \pm 0.173$
& ${\bf 0.854 \pm 0.002}$
& $0.912 \pm 0.002$ \\
\cmidrule(lr){2-8}
&
\multirow{3}{*}{P}
& DPS
& $0.502 \pm 0.005$
& $4.533 \pm 0.094$
& $7.578 \pm 0.414$
& $0.791 \pm 0.005$
& $0.850 \pm 0.003$ \\
&
& DDRM
& $0.380 \pm 0.002$
& $2.464 \pm 0.013$
& $3.067 \pm 0.123$
& $0.843 \pm 0.004$
& $0.923 \pm 0.001$ \\
&
& OpDis-PS ({\bf ours})
& ${\bf 0.341 \pm 0.001}^{\star}$
& ${\bf 2.362 \pm 0.011}^{\star}$
& ${\bf 3.035 \pm 0.188}$
& $0.850 \pm 0.004$
& ${\bf 0.931 \pm 0.000}^{\star}$ \\

\bottomrule
\multicolumn{8}{l}{\footnotesize
RE = relative error; ATE = activation time error; LE = localization error;
sCC = spatial corr. coefficient; tCC = temporal corr. coefficient}
\end{tabular}}
\end{table}

\begin{table}[t]
\centering
\caption{{\bf Quantitative results (pathological data; 30 dB).}
We evaluate all methods on 720 pathological test cases comprising LBBB, RBBB, and myocardial scar under lung-conductivity and subcutaneous-fat discrepancy. Data-driven results are mean $\pm$ standard deviation over five training runs; $\star$ denotes $p<0.05$ for the best versus second-best data-driven method (Appendix~\ref{sec:metric-definitions}).}
\label{tab:pathological-30db}
\resizebox{\textwidth}{!}{%
\begin{tabular}{cclccccc}
\toprule
\multirow{2}{*}{Discrepancy}
& & \multirow{2}{*}{Method}
& \multicolumn{5}{c}{Metric} \\
\cmidrule(lr){4-8}
& & & RE\,$\downarrow$
& ATE [ms]\,$\downarrow$
& LE [mm]\,$\downarrow$
& sCC\,$\uparrow$
& tCC\,$\uparrow$ \\
\midrule

\multirow{8}{*}{\rotatebox[origin=c]{90}{\shortstack{lung\\conductivity}}}
& \multirow{3}{*}{R}
& Tikhonov (spatial)
& $0.695$
& $24.121$
& $46.282$
& $0.627$
& $0.738$ \\
&
& Tikhonov (spatio-temporal)
& $0.677$
& $9.680$
& $15.122$
& $0.664$
& $0.756$ \\
&
& Total Variation
& $0.618$
& $8.573$
& $11.379$
& $0.694$
& $0.762$ \\
\cmidrule(lr){2-8}
&
\multirow{2}{*}{A}
& Transformer
& $0.561 \pm 0.002$
& $7.144 \pm 0.079$
& $7.331 \pm 0.223$
& $0.746 \pm 0.005$
& $0.792 \pm 0.002$ \\
&
& Palette
& $0.517 \pm 0.003$
& $6.404 \pm 0.218$
& $7.988 \pm 0.471$
& $0.786 \pm 0.001$
& $0.821 \pm 0.003$ \\
\cmidrule(lr){2-8}
&
\multirow{3}{*}{P}
& DPS
& $0.618 \pm 0.004$
& $11.226 \pm 0.179$
& $12.245 \pm 0.580$
& $0.722 \pm 0.003$
& $0.738 \pm 0.003$ \\
&
& DDRM
& $0.431 \pm 0.003$
& ${\bf 4.363 \pm 0.116}^{\star}$
& ${\bf 3.938 \pm 0.142}$
& ${\bf 0.837 \pm 0.002}^{\star}$
& ${\bf 0.884 \pm 0.003}^{\star}$ \\
&
& OpDis-PS ({\bf ours})
& ${\bf 0.429 \pm 0.004}^{\star}$
& $4.506 \pm 0.133$
& $4.004 \pm 0.107$
& $0.834 \pm 0.004$
& $0.882 \pm 0.003$ \\

\midrule

\multirow{8}{*}{\rotatebox[origin=c]{90}{\shortstack{subcutaneous\\fat}}}
& \multirow{3}{*}{R}
& Tikhonov (spatial)
& $0.717$
& $24.670$
& $45.018$
& $0.619$
& $0.734$ \\
&
& Tikhonov (spatio-temporal)
& $0.693$
& $9.421$
& $15.365$
& $0.658$
& $0.753$ \\
&
& Total Variation
& $0.636$
& $8.510$
& $10.458$
& $0.692$
& $0.763$ \\
\cmidrule(lr){2-8}
&
\multirow{2}{*}{A}
& Transformer
& $0.565 \pm 0.002$
& $7.108 \pm 0.085$
& $7.362 \pm 0.269$
& $0.746 \pm 0.005$
& $0.792 \pm 0.002$ \\
&
& Palette
& $0.521 \pm 0.003$
& $6.504 \pm 0.239$
& $8.206 \pm 0.693$
& $0.784 \pm 0.002$
& $0.820 \pm 0.003$ \\
\cmidrule(lr){2-8}
&
\multirow{3}{*}{P}
& DPS
& $0.617 \pm 0.004$
& $10.664 \pm 0.168$
& $10.552 \pm 0.335$
& $0.728 \pm 0.003$
& $0.751 \pm 0.003$ \\
&
& DDRM
& $0.444 \pm 0.003$
& ${\bf 4.083 \pm 0.113}^{\star}$
& ${\bf 3.791 \pm 0.145}$
& ${\bf 0.840 \pm 0.002}$
& ${\bf 0.889 \pm 0.002}^{\star}$ \\
&
& OpDis-PS ({\bf ours})
& ${\bf 0.421 \pm 0.003}^{\star}$
& $4.274 \pm 0.121$
& $3.877 \pm 0.105$
& $0.837 \pm 0.004$
& $0.887 \pm 0.003$ \\

\bottomrule
\multicolumn{8}{l}{\footnotesize
RE = relative error; ATE = activation time error; LE = localization error;
sCC = spatial corr. coefficient; tCC = temporal corr. coefficient}
\end{tabular}}
\end{table}

On physiological activation patterns, OpDis-PS achieves the lowest RE and ATE and the highest tCC under both discrepancy sources. Palette attains the highest sCC. DDRM yields the lowest LE under lung-conductivity discrepancy and OpDis-PS under subcutaneous-fat discrepancy. The differences in LE and sCC between the respective best and second-best methods are not statistically significant.

For pathological activation patterns, OpDis-PS retains the lowest RE under both discrepancy sources. DDRM performs best on ATE, LE, sCC, and tCC. The differences in LE under both discrepancy sources and in sCC under subcutaneous-fat discrepancy are not statistically significant.

At 30 dB, the method ordering is largely preserved, but the margins between methods shrink. This narrowing is expected. As measurement noise grows, it increasingly masks the operator discrepancy, so modeling the discrepancy contributes less and the methods converge. The effect is consistent with the noise and discrepancy magnitudes reported in Appendix~\ref{sec:discrepancy-magnitude}.

\subsection{Ablation of the discrepancy model}
\label{sec:ablation-covariance}

We compare the signal-conditional likelihood of OpDis-PS with two marginal approximation-error variants. The first removes signal conditioning and models $\mathbf d_t\sim\mathcal N(\widehat{\boldsymbol\mu}_{\mathrm d},\widehat{\mathbf C}_{\mathrm{dd}})$. We handle the resulting correlated-noise likelihood by whitening as in CARD \citep{Nezakati2026CARD}. The second replaces $\widehat{\mathbf C}_{\mathrm{dd}}$ with the matched-trace isotropic covariance $\widehat{\sigma}^2\mathbf I_{n_{\mathrm B}-1}$, where 
\begin{equation}
    \widehat{\sigma}^2=\frac{\operatorname{tr}(\widehat{\mathbf C}_{\mathrm{dd}})}{n_{\mathrm B}-1},
\end{equation} while retaining $\widehat{\boldsymbol\mu}_{\mathrm d}$. The first comparison assesses the effect of signal conditioning. The second isolates the covariance structure at fixed total discrepancy variance.

Across both discrepancy sources, the automatic Ledoit--Wolf estimator achieves a mean geometry-held-out Gaussian log score comparable to inner-CV-selected fixed shrinkage and higher than that of the matched-trace isotropic covariance.

Tables~\ref{tab:ablation-physiological-40db} and~\ref{tab:ablation-pathological-40db} report these three variants under the same evaluation setup as the main comparison.

\begin{table}[t]
\centering
\caption{{\bf Ablation (physiological data; 40 dB).}
We evaluate the discrepancy-model variants on the 250 physiological test cases under lung-conductivity and subcutaneous-fat discrepancy. Values show the mean and standard deviation over five training runs. $\star$ denotes $p<0.05$ for the best versus second-best variant (two-sided paired $t$-test).}
\label{tab:ablation-physiological-40db}
\resizebox{\textwidth}{!}{%
\begin{tabular}{clccccc}
\toprule
Discrepancy & Method & RE\,$\downarrow$ & ATE [ms]\,$\downarrow$ & LE [mm]\,$\downarrow$ & sCC\,$\uparrow$ & tCC\,$\uparrow$ \\
\midrule
\multirow{3}{*}{\shortstack{lung\\conductivity}}
& OpDis-PS (ours)
    & ${\bf 0.313 \pm 0.001}^{\star}$
    & ${\bf 2.033 \pm 0.008}^{\star}$
    & ${\bf 2.198 \pm 0.146}^{\star}$
    & ${\bf 0.858 \pm 0.003}^{\star}$
    & ${\bf 0.943 \pm 0.000}^{\star}$ \\
& OpDis-PS (marginal)
    & $0.315 \pm 0.001$
    & $2.074 \pm 0.008$
    & $2.312 \pm 0.094$
    & $0.857 \pm 0.003$
    & $0.941 \pm 0.000$ \\
& OpDis-PS (marginal, matched isotropic)
    & $0.365 \pm 0.001$
    & $2.554 \pm 0.009$
    & $3.110 \pm 0.231$
    & $0.818 \pm 0.002$
    & $0.921 \pm 0.001$ \\
\midrule
\multirow{3}{*}{\shortstack{subcutaneous\\fat}}
& OpDis-PS (ours)
    & ${\bf 0.321 \pm 0.001}^{\star}$
    & ${\bf 2.109 \pm 0.005}^{\star}$
    & ${\bf 2.470 \pm 0.090}^{\star}$
    & ${\bf 0.855 \pm 0.004}^{\star}$
    & ${\bf 0.940 \pm 0.000}^{\star}$ \\
& OpDis-PS (marginal)
    & $0.353 \pm 0.002$
    & $2.429 \pm 0.016$
    & $3.642 \pm 0.097$
    & $0.852 \pm 0.004$
    & $0.926 \pm 0.001$ \\
& OpDis-PS (marginal, matched isotropic)
    & $0.425 \pm 0.002$
    & $3.270 \pm 0.030$
    & $4.618 \pm 0.115$
    & $0.790 \pm 0.003$
    & $0.893 \pm 0.001$ \\
\bottomrule
\end{tabular}}
\end{table}

\begin{table}[t]
\centering
\caption{{\bf Ablation (pathological data; 40 dB).}
We evaluate the discrepancy-model variants on the 720 pathological test cases (LBBB, RBBB, and myocardial scar) under lung-conductivity and subcutaneous-fat discrepancy. Values show the mean and standard deviation over five training runs. $\star$ denotes $p<0.05$ for the best versus second-best variant (two-sided paired $t$-test).}
\label{tab:ablation-pathological-40db}
\resizebox{\textwidth}{!}{%
\begin{tabular}{clccccc}
\toprule
Discrepancy & Method & RE\,$\downarrow$ & ATE [ms]\,$\downarrow$ & LE [mm]\,$\downarrow$ & sCC\,$\uparrow$ & tCC\,$\uparrow$ \\
\midrule
\multirow{3}{*}{\shortstack{lung\\conductivity}}
& OpDis-PS (ours)
    & ${\bf 0.393 \pm 0.003}^{\star}$
    & ${\bf 3.621 \pm 0.091}^{\star}$
    & ${\bf 2.927 \pm 0.077}$
    & ${\bf 0.853 \pm 0.003}^{\star}$
    & ${\bf 0.903 \pm 0.002}^{\star}$ \\
& OpDis-PS (marginal)
    & $0.395 \pm 0.003$
    & $3.716 \pm 0.097$
    & $3.078 \pm 0.088$
    & $0.850 \pm 0.003$
    & $0.901 \pm 0.002$ \\
& OpDis-PS (marginal, matched isotropic)
    & $0.438 \pm 0.004$
    & $4.610 \pm 0.140$
    & $4.328 \pm 0.126$
    & $0.819 \pm 0.002$
    & $0.878 \pm 0.003$ \\
\midrule
\multirow{3}{*}{\shortstack{subcutaneous\\fat}}
& OpDis-PS (ours)
    & ${\bf 0.395 \pm 0.003}^{\star}$
    & ${\bf 3.602 \pm 0.091}^{\star}$
    & ${\bf 3.160 \pm 0.086}^{\star}$
    & ${\bf 0.853 \pm 0.003}^{\star}$
    & ${\bf 0.903 \pm 0.002}^{\star}$ \\
& OpDis-PS (marginal)
    & $0.430 \pm 0.004$
    & $4.196 \pm 0.123$
    & $4.113 \pm 0.069$
    & $0.833 \pm 0.004$
    & $0.886 \pm 0.003$ \\
& OpDis-PS (marginal, matched isotropic)
    & $0.514 \pm 0.007$
    & $6.750 \pm 0.391$
    & $6.540 \pm 0.223$
    & $0.771 \pm 0.004$
    & $0.833 \pm 0.006$ \\
\bottomrule
\end{tabular}}
\end{table}

Signal conditioning yields small improvements for lung-conductivity discrepancy and clearer improvements for subcutaneous-fat discrepancy for both datasets. The structured marginal covariance outperforms its matched-isotropic counterpart across all metrics and both discrepancy types.

\subsection{Inference cost}
\label{sec:inference-cost}

\begin{table}[t]
\centering
\caption{{\bf Inference cost.} Median reconstruction time over three repetitions of eight cases.}
\label{tab:inference-cost}
\begin{tabular}{llr}
\toprule
Method & Device & Time \\
\midrule
Tikhonov (spatial) & CPU & $20.12$ ms \\
Tikhonov (spatio-temporal) & CPU & $2.14$ s \\
Total Variation & CPU & $53.91$ s \\
\midrule
Transformer & GPU & $1.79$ ms \\
Palette & GPU & $21.54$ s \\
\midrule
DPS & GPU & $44.62$ s \\
DDRM & GPU & $6.03$ s \\
OpDis-PS (ours) & GPU & $6.04$ s \\
\bottomrule
\end{tabular}
\end{table}

We measure reconstruction time on eight held-out lung-conductivity-discrepancy cases at 20 dB. We use one trained instance per learned model and run all methods sequentially, processing one case at a time. Learned methods use an NVIDIA A100 GPU, while variational methods use 16 CPU threads. Except for Total Variation, whose number of iterations depends on its convergence tolerance, all methods perform a fixed amount of computation per case, so their reconstruction times do not depend on the noise level.

After one excluded warm-up call per method, we reconstruct each case three times. We time complete reconstruction calls with the models already loaded and synchronize CUDA before and after each GPU call. For sampling-based methods, each call includes all 16 draws and their averaging. Table~\ref{tab:inference-cost} reports the median over the resulting 24 measurements.

\section{Simulation details}
\label{sec:simulation-details}

We detail the geometries, electrophysiological simulations, and pathological configurations summarized in Section~\ref{sec:data}.

\subsection{Geometries}

{\bf Hearts.}
For each ventricular mesh, we compute fiber directions with a rule-based method~\citep{Bayer2012}, using helix angles of $80^\circ$ and $-60^\circ$ at the endocardium and epicardium. Ventricular coordinates~\citep{Bayer2018} provide an apico-basal coordinate $Z\in[0,1]$, a transmural coordinate $\rho\in[0,1]$ from endocardium to epicardium, a rotational coordinate $\varphi$, and a ventricle indicator $V$ ($-1$ for the left and $+1$ for the right ventricle). We use them to place stimuli, assign action-potential durations, and define pathological regions.

{\bf Torsos and electrodes.}
We register each heart to the ventricular envelope of the torso CT with a similarity iterative-closest-point fit and accept a pairing only if the epicardium lies inside the torso and does not intersect the lungs. Electrodes lie on a $10\times20$ grid in normalized torso height and angle around the torso centerline, excluding the arm regions. The resulting $176$ positions are the same for every torso, so electrode indices correspond across subjects.

\subsection{Electrophysiology}

We solve the pseudo-bidomain equations in openCARP~\citep{Plank2021openCARP} with the Mitchell--Schaeffer ionic model~\citep{Mitchell2003}. Myocardial conductivities are transversely isotropic with respect to the fiber direction and tuned with openCARP's tuneCV utility on a one-dimensional cable at the mean mesh edge length of $1.13$~mm, targeting conduction velocities of $0.6$ and $0.3$~m/s along and across the fibers~\citep{Gillette2023MedalCare}. A subendocardial layer ($\rho<0.1$, $0.1\leq Z\leq0.9$) with an isotropic conduction velocity of $2$~m/s represents the fast-conducting Purkinje system~\citep{Gillette2023MedalCare}.

{\bf Action-potential durations.}
Following~\citet{Gillette2023MedalCare}, we map the normalized combination $q_Z Z+q_\rho\rho+q_V V$ affinely to $[\mathrm{APD}_{\min},\mathrm{APD}_{\max}]$, with $q_Z\sim\mathcal U[0.9,1]$, $q_\rho\sim\mathcal U[-0.6,0]$, $q_V\sim\mathcal U[0.1,0.15]$, $\mathrm{APD}_{\min}\sim\mathcal U[150,175]$~ms, and $\mathrm{APD}_{\max}\sim\mathcal U[225,250]$~ms. This yields apico-basal, transmural, and interventricular repolarization gradients.

{\bf Activation.}
Five endocardial sites represent the fascicles: anterior, posterior, and septal sites in the left ventricle, and septal and moderator-band sites in the right ventricle. Their positions and radii are sampled from fixed ranges in ventricular coordinates.

{\bf Physiological rounds.}
A physiological round is one draw of the action-potential and activation parameters per heart--torso pair. Training pairs are simulated in three rounds, and validation and test pairs in one.

\subsection{Pathologies}

Table~\ref{tab:pathology-configurations} lists the twelve pathological configurations. Within each pathology, all four configurations use the same 60 held-out test pairs, each with its own draw of the physiological parameters.

{\bf Bundle branch blocks.}
Following~\citet{Gillette2023MedalCare}, LBBB removes the three left-ventricular stimulus sites and RBBB the two right-ventricular sites, so the blocked ventricle is activated through its fast endocardial layer and across the septum. The more severe configurations slow this layer to $1.0$ or $0.8$~m/s or remove it entirely.

{\bf Myocardial scar.}
The scar configurations keep physiological activation and add an electrically inert region, around which the activation front is routed. The region is anchored in the mid-to-basal anterolateral left-ventricular wall, and its size and transmural extent grow together from a subendocardial to a large transmural scar.

\begin{table}[t]
\centering
\caption{{\bf Pathological configurations.}
Severity configurations of the held-out pathologies, each simulated on 60 test pairs. For bundle branch blocks, CV denotes the conduction velocity of the fast endocardial layer of the blocked ventricle; in the complete block, this layer conducts like ordinary myocardium. For scar, $r$ and $\rho_{\max}$ denote the radius and transmural depth of the inert region.}
\label{tab:pathology-configurations}
\begin{tabular}{ccccc}
\toprule
\multirow{2}{*}{Severity} & LBBB & RBBB & \multicolumn{2}{c}{Scar} \\
\cmidrule(lr){2-2}\cmidrule(lr){3-3}\cmidrule(lr){4-5}
& CV [m/s] & CV [m/s] & $r$ [mm] & $\rho_{\max}$ \\
\midrule
1 & $2.0$ & $2.0$ & $10$ & $0.4$ \\
2 & $1.0$ & $1.0$ & $15$ & $0.7$ \\
3 & $0.8$ & $0.8$ & $20$ & $1.0$ \\
4 & complete & complete & $22$ & $1.0$ \\
\bottomrule
\end{tabular}
\end{table}

\end{document}